\documentclass[sigconf]{acmart}
\AtBeginDocument{%
  }

\copyrightyear{2025}
\acmYear{2025}
\setcopyright{acmlicensed}\acmConference[KDD '25]{Proceedings of the 31st ACM SIGKDD Conference on Knowledge Discovery and Data Mining V.2}{August 3--7, 2025}{Toronto, ON, Canada}
\acmBooktitle{Proceedings of the 31st ACM SIGKDD Conference on Knowledge Discovery and Data Mining V.2 (KDD '25), August 3--7, 2025, Toronto, ON, Canada}
\acmDOI{10.1145/3711896.3737160}
\acmISBN{979-8-4007-1454-2/2025/08}

\usepackage{multirow}
\usepackage{makecell}
\usepackage[caption=false,font=footnotesize]{subfig}

\begin{document}

\settopmatter{printacmref=true} 

\title{Toward End-to-End Bearing Fault Diagnosis for Industrial Scenarios with Spiking Neural Networks}

\author{Lin Zuo}
\authornote{Corresponding author: Lin Zuo.\\The code is available at https://github.com/yqding326/MRA-SNN.}
\email{linzuo@uestc.edu.cn}
\affiliation{%
  \institution{School of Information and Software Engineering\\University of Electronic Science and Technology of China}
  \city{Chengdu}
  \state{Sichuan}
  \country{China}}

\author{Yongqi Ding}
\email{yqding@std.uestc.edu.cn}
\affiliation{%
  \institution{School of Information and Software Engineering\\University of Electronic Science and Technology of China}
  \city{Chengdu}
  \state{Sichuan}
  \country{China}
}

\author{Mengmeng Jing}
\email{jingmeng1992@gmail.com}
\affiliation{%
  \institution{School of Information and Software Engineering\\University of Electronic Science and Technology of China}
  \city{Chengdu}
  \state{Sichuan}
  \country{China}
}

\author{Kunshan Yang}
\email{ksyang@std.uestc.edu.cn}
\affiliation{%
  \institution{School of Information and Software Engineering\\University of Electronic Science and Technology of China}
  \city{Chengdu}
  \state{Sichuan}
  \country{China}
}

\author{Biao Chen}
\email{chenbiao@std.uestc.edu.cn}
\affiliation{%
  \institution{School of Information and Software Engineering\\University of Electronic Science and Technology of China}
  \city{Chengdu}
  \state{Sichuan}
  \country{China}
}

\author{Yunqian Yu}
\email{yuyunqianyyz@gmail.com}
\affiliation{%
  \institution{School of Information and Software Engineering\\University of Electronic Science and Technology of China}
  \city{Chengdu}
  \state{Sichuan}
  \country{China}
}

\renewcommand{\shortauthors}{Lin Zuo et al.}

\begin{abstract}
This paper explores the application of spiking neural networks (SNNs), known for their low-power binary spikes, to bearing fault diagnosis, bridging the gap between high-performance AI algorithms and real-world industrial scenarios. In particular, we identify two key limitations of existing SNN fault diagnosis methods: inadequate encoding capacity that necessitates cumbersome data preprocessing, and non-spike-oriented architectures that constrain the performance of SNNs. To alleviate these problems, we propose a Multi-scale Residual Attention SNN (MRA-SNN) to simultaneously improve the efficiency, performance, and robustness of SNN methods. By incorporating a lightweight attention mechanism, we have designed a multi-scale attention encoding module to extract multiscale fault features from vibration signals and encode them as spatio-temporal spikes, eliminating the need for complicated preprocessing. Then, the spike residual attention block extracts high-dimensional fault features and enhances the expressiveness of sparse spikes with the attention mechanism for end-to-end diagnosis. In addition, the performance and robustness of MRA-SNN is further enhanced by introducing the lightweight attention mechanism within the spiking neurons to simulate the biological dendritic filtering effect. Extensive experiments on MFPT, JNU, Bearing, and Gearbox benchmark datasets demonstrate that MRA-SNN significantly outperforms existing methods in terms of accuracy, energy consumption, and noise robustness, and is more feasible for deployment in real-world industrial scenarios.

\end{abstract}



\maketitle

\section{Introduction}

Spiking neural networks (SNNs), which mimic the information transmission mechanism of biological neural systems, have attracted considerable attention for their low-energy paradigm~\cite{9543525,roy2019towards}. Specifically, SNNs transmit information via discrete 0-1 spikes. Spiking neurons are silenced for 0-valued input spikes and only need to perform accumulation (AC) operations for 1-valued spikes (event-driven)~\cite{roy2019towards}. In contrast, the current widely used artificial neural networks (ANNs) have intensive multiply-accumulate (MAC) operations. In the typical case of a 32-bit floating-point implementation in 45nm technology~\cite{6757323}, the AC operation consumes $0.9pJ$ of power, while the MAC operation requires $4.6pJ$, more than five times that of the AC operation. Thus, even when compared to optimized lightweight ANNs, SNNs still have significant power consumption advantages, making them preferable for deployment in energy- and latency-sensitive edge devices. For example, Spiking-YOLO~\cite{Spiking_YOLO} consumes 280 times less power than ANN-Tiny YOLO, and Speck~\cite{yao2024spike} requires only 0.7 mW of power to perform typical vision tasks.

Benefiting from the advantages of low power consumption and high bionicity, SNNs have been used widely in computer vision, reinforcement learning and other fields~\cite{MSResNet,SSNN,qin2023low}. However, SNNs are rarely explored for industrial scenarios related to real-world applications. Industrial tasks have a huge impact on the normal operation of equipment and even the safety of personnel, such as typical bearing fault diagnosis, which often requires fast and robust algorithmic support~\cite{10555174}. Existing fault diagnosis methods typically use ANNs, which provide decent results but still struggle with high latency and energy consumption~\cite{DRSN,wang2023bearing}. To overcome this energy-performance dilemma, SNNs have been introduced into bearing fault diagnosis with promising results~\cite{zuosnn,DSRSN,MLR-SNN}. Unfortunately, these SNN methods are either constrained to shallow fully connected forms~\cite{zuosnn,MLSNN} or residual network architectures~\cite{MLR-SNN,DSRSN} like ANNs, which do not consider spike properties and suffer from limited diagnostic performance and efficiency. Therefore, it remains necessary to further explore efficient and high-performance SNN fault diagnosis methods for real industrial environments to facilitate the deployment of next-generation AI algorithms.

In this paper, we first identify two key factors that limit the diagnostic performance of SNNs (See Section~\ref{problem_analysis} for details): (1) inadequate spike encoding capacity requires additional data preprocessing, and (2) network architectures that do not account for spike characteristics, resulting in suboptimal performance. To this end, we propose a Multi-scale Residual-Attention SNN (MRA-SNN): extracting multi-scale features in the data and adaptively fusing them for spike encoding with the attention mechanism, thus eliminating tedious data preprocessing and dramatically improving the diagnostic efficiency; and rectifying the high-dimensional residual features with the attention mechanism for the increasingly sparse spikes to improve the representation performance of the SNN. Moreover, inspired by the filtering of input currents by dendrites in biological neurons~\cite{SPRUSTON1994161,Magee2000DendriticIO}, we introduce the attention mechanism in spiking neurons to mimic this dendritic filtering effect. This enhances the discriminative ability of the spiking neurons~\cite{PASNN}, which further improves the overall performance and robustness of the MRA-SNN for accurate fault diagnosis under noisy interference. The attention mechanism we use is lightweight and involves only single-channel 1D convolutions for channel-spatial attention, making MRA-SNN concise and effective. Extensive experiments on the challenging MFPT, JNU, Bearing, and Gearbox benchmarks demonstrate the superior performance of MRA-SNN. Compared to other existing SNN fault diagnosis methods, the lightweight MRA-SNN shows better performance in both normal and noisy environments, even surpassing ANN methods. In summary, the main contributions of this paper are as follows:
\begin{enumerate}
\item We propose MRA-SNN for bearing fault diagnosis, with a multi-scale attention encoding module to convert vibration signals into spikes, thus eliminating cumbersome data preprocessing, and a spike residual attention block to enhance the representational capability of the network.
\item We introduces the lightweight attention mechanism in spiking neurons to simulate the filtering behavior of biological dendrites. This enhances the bionic and discriminative properties of the spiking neurons, as well as the fault diagnosis performance and noise robustness of the MRA-SNN.
\item Extensive experiments on the MFPT, JNU, Bearing, and Gearbox benchmarks confirm the effectiveness of our method, which achieves superior performance in both normal and noisy environments with significantly lower energy consumption compared to existing methods.
\end{enumerate}

\section{Related Work and Background} 
\subsection{Spiking Neural Network}
\label{snn_intro}
As the third generation of neural networks, SNNs provide an ultra-low power computing paradigm by eliminating MAC operations in ANNs through binary spike communication. The low power advantage makes SNNs extremely attractive in real-world scenarios and widely used in various fields. In computer vision, for example, SNNs are used for object recognition~\cite{STBP,PLIF}, detection~\cite{Spiking_YOLO,Su_2023_ICCV}, and tracking~\cite{luo2021siamsnn}. For reinforcement learning, SNNs have been able to perform game and motion control~\cite{qin2023low,tang2021deep}. With the advent of the large model era, the spiking Large Language Model has also achieved impressive results~\cite{bal2024spikingbert,zhu2023spikegpt}. In terms of hardware devices, Kim et al.~\cite{Spiking_YOLO} uses the SNN for object detection, which consumes 280 times less power than its ANN counterpart, and Yao et al.~\cite{yao2024spike} uses the SNN for typical vision tasks, which consumes only 0.7 mW. In this paper, we aim to push the SNN to the industry to better utilize its low power consumption and high efficiency to advance the task of mechanical bearing fault diagnosis.
\subsection{Fault Diagnosis}
Fault diagnosis aims to detect device faults according to the 1-D vibration signals collected by the device side sensors. Early methods built sophisticated device-dependent mathematical-physical models, but were not applicable to increasingly complex mechanical systems~\cite{li2021intelligent}. Currently, data-driven ANN-based methods capable of adaptively learning and diagnosing from large amounts of historical data are the most popular methods. For example, Zhao et al.~\cite{DRSN} proposed deep residual shrinkage network (DRSN) for robust fault diagnosis based on convolutional neural networks (CNNs). Chen et al.~\cite{chen2021bearing} combined CNN and long short-term memory (LSTM) to extract fault-related features from raw vibration signals. However, high-performance ANNs demand huge energy consumption~\cite{feldmann2019all}, which makes these methods hardly feasible for practical edge devices. To this end, methods based on distillation~\cite{10496469} and lightweight architectures~\cite{LEFE-Net} have been proposed to reduce power consumption, but due to the inherent nature of ANNs, they still suffer from severe power consumption challenges. Therefore, it is worth exploring emerging computing paradigms to get out of this dilemma.
\subsection{SNNs in Fault Diagnosis}
\label{problem_analysis}
Previous work has introduced SNNs to the field of fault diagnosis with quite impressive effects. Zuo et al.~\cite{zuosnn} used Local Mean Decomposition (LMD) to extract features from vibration signals and then a single-layer SNN for bearing fault diagnosis. Wang et al.~\cite{ISNN} proposed an improved SNN for intershaft bearing fault diagnosis using short-time Fourier transform (STFT)-Norm-LIF coding and simplifying the backpropagation process of spiking neurons. Based on the probabilistic transmission mechanism, Zuo et al.~\cite{MLSNN} use a multilayer SNN, which outperforms multilayer ANNs and has great transparency. Xu et al.~\cite{DSRSN} proposed deep spiking residual shrinkage network (DSRSN), which achieves robust fault diagnosis under noise interference by using the attention mechanism and soft thresholding. SNN fault diagnosis methods have also been extended to the fault diagnosis of devices other than bearings. Wang et al.~\cite{MLR-SNN} proposed membrane learnable residual SNN (MLR-SNN) for fault diagnosis of sensors in autonomous vehicles. These works confirm the potential of SNNs for fault diagnosis, but still suffer from several serious challenges:
\begin{itemize}
\item Heavy data preprocessing. It is difficult to extract fault features from non-smooth and non-linear vibration signals by directly using SNNs, so the existing methods use LMD~\cite{MLSNN,zuosnn} or STFT~\cite{ISNN} to extract time-frequency features before using SNNs for fault diagnosis. The pre-processing of vibration signals limits the diagnostic efficiency and makes it almost impossible to diagnose faults on-line in real time.
\item Non-spike oriented architecture. Existing methods directly use fully connected or ResNet architectures for ANNs and lack the exploration of architectures that incorporate spike characteristics. Effective architectures that can extract more expressive features consider spike characteristics are imperative to be explored to achieve superior performance.
\end{itemize}
To address these challenges, this paper optimizes the spike encoding, network architecture, and neuron model to eliminate cumbersome data preprocessing and improve the efficiency and performance of the SNN.

\begin{figure*}[!t]
\centering
\includegraphics[width=6.8in]{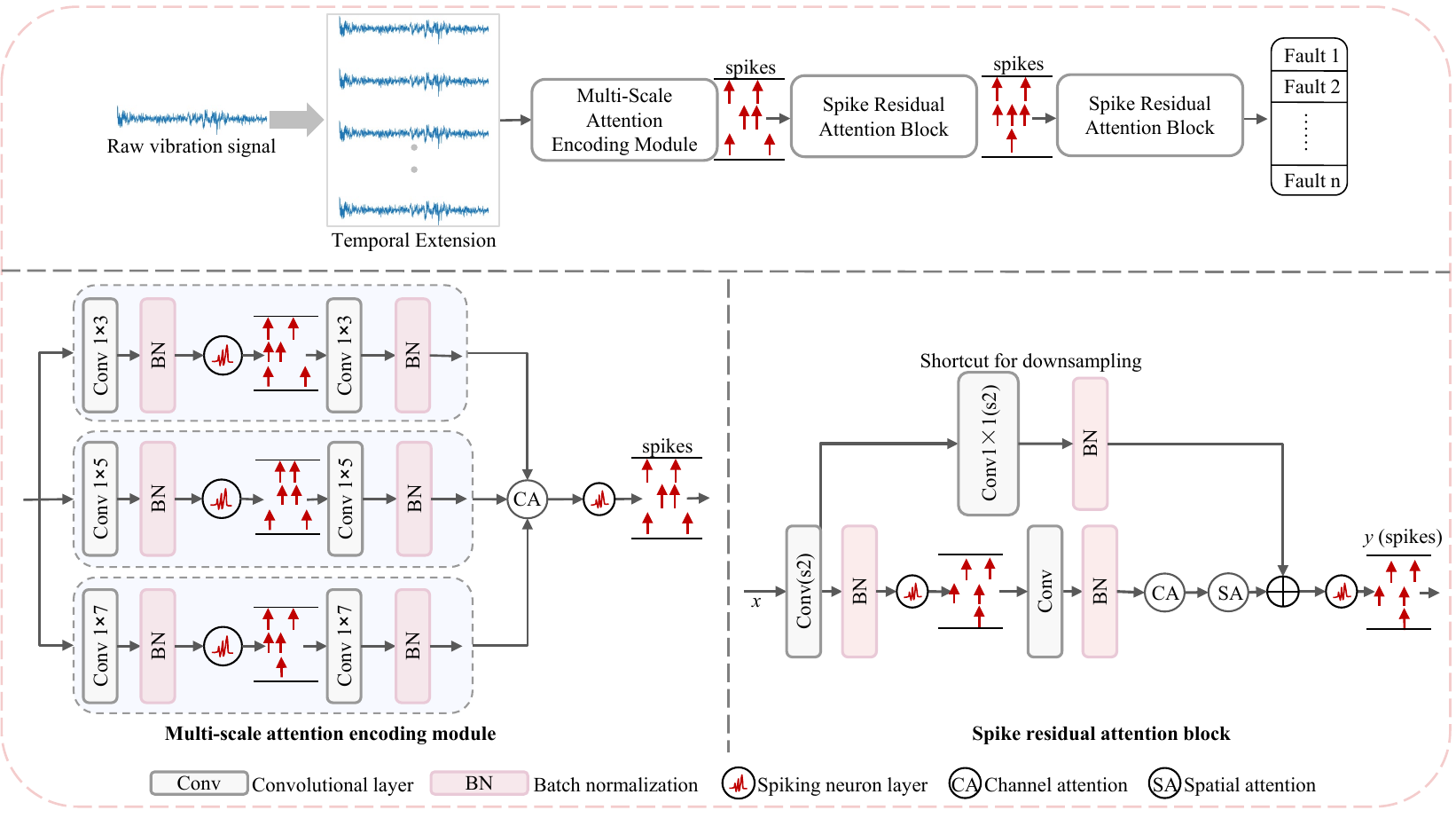}
\caption{Overview of the MRA-SNN framework. The MRA-SNN consists of a multi-scale attention encoding module and two spike residual attention blocks, with a fully connected layer used to classify fault types. The raw vibration signal after temporal extension are fed as input to the MRA-SNN at multiple timesteps without additional data preprocessing.}
\label{overview}
\vskip -0.1in
\end{figure*}

\section{Preliminary} 
\label{preliminary}
This section describes the preliminaries of SNNs, including the dynamics of spiking neurons and the SNN training method used in this work.
\subsection{Spiking Neuron}
Spiking neurons distinguish SNNs from ANNs. Unlike neurons in ANNs such as Rectified Linear Unit (ReLU), spiking neurons model the information transmission mechanism of biological neurons with complicated internal dynamics. Spiking neurons iteratively experienced the process of charging, firing spikes, and resetting membrane potential over time. 

At timestep $t$, the spiking neuron receives the input current $I$ transmitted from the previous layer of neurons and charges the membrane potential $H$ by incorporating it. For the most commonly used leaky integrate-and-fire (LIF)~\cite{STBP} neurons, whose membrane potential leaks over timestep:
\begin{equation}
H_{i}^{l}(t)=\left(1-\frac{1}{\tau}\right) U_{i}^{l}(t-1)+I_{i}^{l}(t),
\label{eq2}
\end{equation}
where $U$ is the membrane potential after resetting at the previous timestep; superscript $l$ and subscript $i$ denote the $i$-th neuron in layer $l$. $\tau$ is the membrane potential constant that controls the leakage rate.

After charging the membrane potential, a spike is generated once the membrane potential reaches the firing threshold $\vartheta$:
\begin{equation}
S_{i}^{l}(t) = \Theta(H_{i}^{l}(t)-\vartheta),
\label{eq3}
\end{equation}
where $\Theta(\cdot)$ denotes the Heaviside step function:
\begin{equation}
\Theta(x)=\left\{
\begin{array}{cl}
1,\quad x \ge 0 \\
0,\quad x < 0 \\
\end{array}.
\right.
\label{eq4}
\end{equation}

After the spike is fired, the spiking neuron resets the membrane potential $U$. This paper uses the soft reset to reduce the membrane potential by a magnitude of the threshold:
\begin{equation}
U_{i}^{l}(t) = r(H_{i}^{l}(t),S_{i}^{l}(t))=H_{i}^{l}(t)-S_{i}^{l}(t)\vartheta.
\label{eq5}
\end{equation}

\subsection{Surrogate Gradient Training}
The spike activity is discontinuous and non-differentiable due to the Heaviside step function, which prevents the back-propagation (BP) algorithm from being used directly to optimize SNNs. To obtain high performance SNNs, the surrogate gradient-based method generates spikes during forward propagation using the Heaviside step function, and replaces the Heaviside step function during backward propagation with a predefined surrogate function $h(\cdot)$ to calculate the gradient. The smooth surrogate functions enable feasible optimization of parameters in SNNs based on the BP algorithm. Specifically, the gradient of the spike w.r.t. the membrane potential can be calculated as:
\begin{equation}
\frac{\partial S_{i}^{l}(t)}{\partial H_{i}^{l}(t)} \approx \frac{\partial h(H_{i}^{l}(t), \vartheta)}{\partial H_{i}^{l}(t)}.
\label{eq6}
\end{equation}

This work uses the rectangular surrogate function~\cite{STBP}:
\begin{equation}
h(H_{i}^{l}(t), \vartheta) = \frac{1}{a} sign\left(\left|H_{i}^{l}(t)-\vartheta\right|<\frac{a}{2}\right),
\label{eq7}
\end{equation}
where $a = 1$ is a hyperparameter that controls the shape of the rectangular function.

\section{Methodology} 
\label{Methodology}

The overall schematic of the MRA-SNN is shown in Fig.~\ref{overview}. The multi-scale attention encoding module encodes the raw vibration signals directly into spikes without the need for heavy data preprocessing to extract time-frequency domain features. The subsequent two spike residual attention blocks extract fault-related high-dimensional features taking advantage of residual learning and rectify the sparse spike residual information through the attention mechanism. Finally, the fully connected layer is used to classify fault types for end-to-end bearing fault diagnosis. Note that since the SNN runs over multiple timesteps (denoted by $T$), the raw vibration signal is temporally extended into $T$ identical signals that are repeatedly input to the MRA-SNN at each timestep. This temporal extension does not affect the efficiency because it does not involve data computation or time-frequency domain feature extraction. The details of the multi-scale attention encoding module, the spike residual attention block, and the attention spiking neuron are described in detail below.

\subsection{Multi-Scale Attention Encoding Module}
For bearing fault diagnosis, it is crucial to extract critical information from non-smooth, non-linear vibration signals and encode it as spikes. Previous methods use LMD~\cite{MLSNN,zuosnn} or STFT~\cite{ISNN} to preprocess the vibration signals and then a simple SNN to classify the faults, which greatly affects the diagnostic efficiency. In order to avoid the heavy preprocessing, the multi-scale attention encoding module was specially designed in this paper to extract key features from the raw vibration signals and encode them into spikes.

The schematic of the multi-scale attention encoding module is shown in the bottom left of Fig.~\ref{overview}. Three convolution pathways with $1 \times 3$, $1 \times 5$, and $1 \times 7$ convolutional kernels are available for extracting fault features at different scales. The features extracted from multiple scales are more comprehensive than vanilla single-scale SNNs and model the multi-level structure of the biological cortex~\cite{4069258}, providing a basis for accurate fault diagnosis. The convolved features are converted into input current $I$ through the Batch Normalization (BN)~\cite{BN} layer to be transmitted into the spiking neuron, which consequently generates spike sequences. This couples convolution and spiking together, preserving the energy efficiency benefits of SNNs and enabling deployment on neuromorphic chips~\cite{9900453}.

For the second BN layer in each convolution pathway, the current $I$ it generates is not passed directly to the spiking neuron. This is because if all three pathways generate spikes, the fused output becomes an analog value (spikes are added directly or weighted), thus losing the low-power characteristic of 0-1 spikes. Instead, we first used channel attention to selectively focus on the currents of the three pathways on a channel-wise basis to distinguish the importance of different scales of information. The additive fusion of the filtered currents is then fed to the spiking neurons to accumulate membrane potential and fire spikes. In this way, the efficient 0-1 spike output is maintained, while effective fusion of multi-scale information is achieved. The visualization of the multi-scale pathways and fused spikes is shown in \textbf{Appendix~\ref{msfvis}} to more clearly illustrate the extracted multi-scale feature information.

Let $X$ denote the input raw vibration signal, the process of generating multi-scale currents can be formulated as:
\begin{equation}
\label{eq8}
I_3 = bn(conv_{1 \times 3}(pool(sn(bn(conv_{1 \times 3}(X)))))),
\end{equation}
\begin{equation}
\label{eq9}
I_5 = bn(conv_{1 \times 5}(pool(sn(bn(conv_{1 \times 5}(X)))))),
\end{equation}
\begin{equation}
\label{eq10}
I_7 = bn(conv_{1 \times 7}(pool(sn(bn(conv_{1 \times 7}(X)))))),
\end{equation}
where $conv(\cdot)$ denotes the convolution layer, $bn(\cdot)$ represents the BN layer, and $sn(\cdot)$ is the spiking neuron layer. To reduce the size of the features as well as the computational overhead, the spike maps generated by the first spiking neuron layer were downsampled using average pooling, denoted by $pool(\cdot)$, with stride set to 2.

The fusion of multi-scale currents to accumulate membrane potential and generate spikes can be formulated as:
\begin{equation}
\label{eq11}
S = sn(ca(I_3; I_5; I_7) \cdot (I_3; I_5; I_7)),
\end{equation}
where $(a;b;c)$ denotes the concatenation operation along the channel dimension and $ca(\cdot)$ is the channel attention, which will be detailed in Section \ref{Attention}.

\begin{figure*}[!t]
\centering
\includegraphics[width=6.8in]{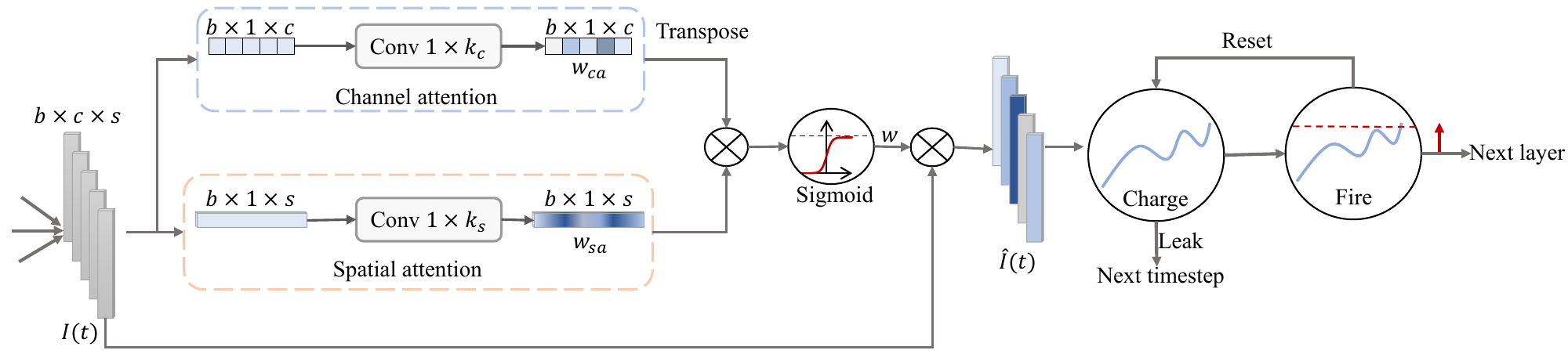}
\vskip -0.1in
\caption{The schematic of the proposed lightweight attention spiking neuron. Channel attention and spatial attention synergistically filter the input current to model dendrites in the biological neural system.}
\label{asn}
\vskip -0.12in
\end{figure*}

\subsection{Spike Residual Attention Block}
Residual learning~\cite{ResNet} effectively mitigates the information and gradient vanishing problem in deep neural networks, preventing performance degradation. Based on this, we construct spike residual attention blocks for extracting abstract fault features in MRA-SNN and preventing information vanishing. Considering the large length of the bearing vibration signal, we need to continuously reduce the feature map to decrease the computational cost. This prevents the identity connections commonly used in ANNs from being used in the spike residual attention block. To do this, each block downsamples the input feature map through the first convolutional layer on the residual and the shortcut branch, and accumulates the sum of the two pathways. For the implementation, the stride of the first convolution layer on the residual and shortcut pathways is set to 2, as shown in the bottom right of Fig.~\ref{overview}.

On the other hand, the spikes in SNNs become sparser as the layer deepens, so it is necessary to improve the expressiveness of sparse spikes. Therefore, we refine the features extracted from the residual pathways using joint channel-spatial attention to amplify/suppress critical/redundant features. Both channel and spatial attention are implemented by one-dimensional convolution and the sigmoid function, described in detail in Section \ref{Attention}, with only negligible computational overhead. This feature refinement is used for the output of the BN layer, which can be regarded as a modulation of the input current to the spiking neurons, to some extent modeling the information filtering mechanism of the biological nervous system~\cite{SPRUSTON1994161,Magee2000DendriticIO}. This practice is somewhat similar to~\cite{10032591}, but we do not adjust the membrane potential of the spiking neuron, thus eliminating the need to couple attention to the neuron model, and is more conducive to deployment on neuromorphic chips~\cite{9900453}.

Note that both the residual and shortcut pathways generate analog value outputs. Similar to the encoding module, the sum of the outputs of these two pathways is used as the input current to the spiking neuron, which then fires the spike. Therefore, the spike residual attention block outputs discrete 0-1 spikes, maintaining the low energy consumption characteristic of SNNs.

Without loss of generality, let the input to the spike residual block be $x$, the residual pathway can be formulated as:
\begin{equation}
\label{eq12}
I_{residual} = bn(conv(sn(bn(conv_{s2}(x))))),
\end{equation}
where $s2$ is the convolution stride of 2 for downsampling. The shortcut pathway can be formulated as:
\begin{equation}
\label{eq13}
I_{shortcut} = bn(conv^{1\times1}_{s2}(x)),
\end{equation}
where $1 \times 1$ is the convolution kernel size. The output spikes $y$ of the spike residual block can be calculated as:
\begin{equation}
\label{eq14}
y = sn(sa(ca(I_{residual})) + I_{shortcut}),
\end{equation}
where $ca(\cdot)$ and $sa(\cdot)$ are channel and spatial attention, respectively, as detailed in Section \ref{Attention}.

\subsection{Attention Spiking Neuron}
\label{Attention}
Spiking neurons simulate the information transmission mechanism and internal dynamics of biological neurons. Theoretically, the higher the bionicity of the spiking neuron, the more ingenious the internal dynamics and the greater the performance~\cite{GLIF}. However, highly bio-characteristic neurons are challenging to implement in computing platforms. Majority of existing SNNs employ simple LIF~\cite{zuosnn,STBP} neurons or their parameterized variants~\cite{MLR-SNN,PLIF}, which limits the performance of SNNs. Inspired by the filtering of information by dendrites in biological neurons~\cite{SPRUSTON1994161,Magee2000DendriticIO}, this work proposes the attention spiking neuron to model the dendrite with a lightweight channel-spatial attention mechanism. This significantly improves the bionicity and discrimination of spiking neurons with negligible parameter overhead.

Specifically, the attention mechanism is used in the process of charging the membrane potential of a spiking neuron to discriminate information in the input current. The charging process of a spiking neuron can be reformulated as:
\begin{equation}
H_{i}^{l}(t) = f(U_{i}^{l}(t-1),\hat{I}_{i}^{l}(t)),
\label{eq15}
\end{equation}
where $\hat{I}_{i}^{l}(t)$ is the input current filtered by the attention mechanism, expressed as:
\begin{equation}
\hat{I}_{i}^{l}(t) = f_{att}(I_{i}^{l}(t)),
\label{eq16}
\end{equation}
where $ f_{att}(\cdot)$ denotes the attention mechanism. This is similar in form to the attention discrimination mechanism (ADM) in~\cite{PASNN}. However, ADM uses a vanilla convolution layer and a sigmoid function as its attention mechanism. This work, on the other hand, employs a lightweight channel-spatial attention mechanism with less parameter overhead and superior performance.

\begin{table*}[!t]
 \centering
 \caption{Comparative results with other fault diagnosis methods on four benchmark\label{com_result}}
 \vskip -0.12in
 \begin{tabular}{cc|cccc|cc}
  \hline
 \multirow{2}{*}{Method} & \multirow{2}{*}{Type} & \multicolumn{4}{c}{Accuracy $\pm$ std (\%)} & \multirow{2}{*}{Param} & \multirow{2}{*}{Energy($pJ$)}\\
  & & MFPT & JNU & Bearing & Gearbox\\ 
  \hline
  LEFE-Net~\cite{LEFE-Net} & Lightweight ANN & 91.08$\pm$ 0.84 & 93.62$\pm$ 0.47 & 98.63$\pm$ 1.32 & 99.90$\pm$ 0.06 & 0.15M & 0.61G\\
  Distillation~\cite{10496469} & Lightweight ANN & 91.71$\pm$ 0.68 & 91.53$\pm$ 0.50 &- & -& 0.91M & 1.27G\\
  LiConvFormer~\cite{LiConvFormer} & ANN Transformer & 86.35$\pm$ 1.84 & 88.18$\pm$ 0.93 & 98.35$\pm$ 0.73 & 99.68$\pm$ 0.19 & 1.27M & 0.24G\\
  DCA-BiGRU~\cite{ZHANG2022110242} & ANN GRU & 86.31 $\pm$ 2.82 & 82.56 $\pm$ 1.14 &- & - & 3.23M & -\\
  SE-Net~\cite{SE} & ANN Attention & 87.38 $\pm$ 2.61 & 93.45 $\pm$ 1.07 & 98.96 $\pm$ 0.30 & 99.70 $\pm$ 0.08 & 5.24M & - \\
  CBAM~\cite{CBAM} & ANN Attention & 88.52 $\pm$ 0.67 & 93.82 $\pm$ 0.19 & 99.33 $\pm$ 0.38 & 99.74 $\pm$ 0.14 & 5.33M & - \\
  Attention SNN~\cite{10032591} & SNN Attention\textbf{} & 87.68 $\pm$ 0.66 & 93.93 $\pm$ 0.42 &- & - & 3.94M & -\\
  ResNet~\cite{ResNet} & ANN & 87.91 $\pm$ 2.68 & 91.53 $\pm$ 0.74 &- & - & 3.85M & 3.24G\\
  DRSN~\cite{DRSN} & ANN & 88.95 $\pm$ 1.52 & 93.06 $\pm$ 0.52 & 98.14 $\pm$ 1.46 & 99.84 $\pm$ 0.10 & 5.24M & 3.22G\\
  Spiking ResNet & SNN & 85.59 $\pm$ 0.77 & 85.38 $\pm$ 2.33 & 97.14$\pm$ 0.69 & 98.80 $\pm$ 0.35 & 3.85M & 0.50G\\
  DSRSN~\cite{DSRSN} & SNN & 86.13 $\pm$ 1.87 & 92.15 $\pm$ 0.85 & 98.16 $\pm$ 0.75 & 99.46 $\pm$ 0.31 & 5.24M & 0.44G\\
  MLR-SNN~\cite{MLR-SNN} & SNN & 88.35 $\pm$ 1.22 & 86.11 $\pm$ 1.08 &  97.52 $\pm$ 0.21 & 98.22 $\pm$ 0.58 & 3.85M & 0.44G\\
  MS-ResNet~\cite{MSResNet} & SNN & 86.73 $\pm$ 0.72 & 93.41 $\pm$ 1.35 &- & -  & 3.85M & 0.52G\\
  \hline
  \textbf{MRA-SNN (Ours)}  & \textbf{SNN} & \textbf{94.57} $\pm$ 0.82 & \textbf{94.18} $\pm$ 0.29 & \textbf{99.36} $\pm$ 0.21 & \textbf{100.0} $\pm$ 0.00 & \textbf{1.75M} & \textbf{0.05G}\\
  \hline
 \end{tabular}
\vskip -0.1in
\end{table*}

As shown in Fig.~\ref{asn}, the attention mechanism in the proposed attention spiking neuron is composed of two elements: channel attention and spatial attention. For channel attention, the input current is globally averaged in the spatial dimension, and then the channel-wise attention scores are calculated adaptively in the channel dimension using a one-dimensional convolution. This was inspired by~\cite{eca}, and the number of additional parameters required is only the convolution kernel size. The spatial attention is similar to the channel attention where the input current is globally averaged in the channel dimension, and then the attention scores are adaptively calculated in the spatial dimension using another one-dimensional convolution. The element-wise attention weights for filtering the input current are obtained by the product of the channel attention score and the spatial attention score and the sigmoid function. Assuming a one-dimensional convolution of size $k$, channel attention and spatial attention need only $2 \times k$ additional parameters. In this case, the ADM~\cite{PASNN} with vanilla convolution requires $c\times c\times k$ additional parameters, where $c$ is the numbers of the channel.

In this paper, attention filtering is coupled with the internal dynamics of LIF neurons. Let $I^{l}(t)\in \mathbb{R}^{b \times c \times s}$ be the input current, where $b$ denotes the batch size, $c$ is the number of channels, and $s$ indicates the length of the spatial dimension. Global average pooling of $I^{l}(t)$ in spatial and channel dimensions yields $AVG_{c}^{l}(t) \in \mathbb{R}^{b \times c \times 1}$ and $AVG_{s}^{l}(t) \in \mathbb{R}^{b \times 1 \times s}$. To enable the one-dimensional convolution operation, $AVG_{c}^{l}(t)$ is transposed to $\hat{AVG}_{c}^{l}(t) \in \mathbb{R}^{b \times 1 \times c}$. Then one-dimensional convolution is applied to obtain the channel attention score $w_{ca}$ and the spatial attention score $w_{sa}$:
\begin{equation}
w_{ca} = conv^{1\times k_c}(\hat{AVG}_{c}^{l}(t)),
\label{eq17}
\end{equation}
\begin{equation}
w_{sa} = conv^{1\times k_s}(AVG_{s}^{l}(t)),
\label{eq18}
\end{equation}
where $k_c$ and $k_s$ denote the size of the convolution kernel for channel attention and spatial attention, respectively. In this work, $k_s$ is set to 7 and $k_c$ follows~\cite{eca}: $k_c=|\frac{\log _{2}(c)}{2}+\frac{1}{2}|_{odd}$. 

Then, $w_{ca}$ is transposed to $\hat{w}_{ca} \in \mathbb{R}^{b \times c \times 1}$ to obtain the channel-wise attention score. The element-wise attention weights $w \in \mathbb{R}^{b \times c \times s}$ are calculated as:
\begin{equation}
w = \sigma (\hat{w}_{ca} \odot w_{sa}),
\label{eq19}
\end{equation}
where $\sigma(\cdot)$ is the sigmoid function and $\odot$ denotes the product with the broadcast mechanism. The filtered input current $\hat{I}^{l}(t)$ is:
\begin{equation}
\hat{I}^{l}(t) = f_{att}(I^{l}(t))= w \cdot I^{l}(t).
\label{eq16}
\end{equation}
The filtered current $\hat{I}^{l}(t)$ replaces the original current $I^{l}(t)$, accumulating membrane potential and firing spikes based on the dynamics of the LIF neurons.

\section{Experiments}
\label{Experiment}
Our experiments are based on the PyTorch package, running on an Ubuntu system with an NVIDIA TITAN RTX GPU. All models were trained for 100 epochs using the Adam optimizer. The initial learning rate was 0.01, scaled down to 0.1 times the previous rate every 30 epochs. The batch size is 64. For spiking neurons, $\tau = 2.0$ and threshold $\vartheta = 1.0$, and timestep of 4 if not specified. All experiments were repeated five times with different random seeds, and the average accuracy and standard deviation were reported.

To validate the effectiveness of the proposed method, we conduct experiments on four fault diagnosis datasets, namely MFPT~\cite{MFPT}, JNU~\cite{JNU}, Bearing and Gearbox~\cite{Gearbox}. In these datasets, the fault classes were divided into 15, 12, 10, and 10 classes, and each sample was intercepted with 1024 length vibration signals. See \textbf{Appendix~\ref{exp_detail}} for experimental details.

In addition to diagnostic accuracy and standard deviation, we also analyzed the power consumption of various models. The floating-point operations of the ANN induce MAC operations, so its power consumption is the total power consumption of MAC operations, while the SNN has only 1-valued spikes that induce AC operations, and its power consumption depends on the number of spikes. See \textbf{Appendix~\ref{EnergyAna}} for detailed power consumption calculations.

\subsection{Evaluation and Comparison}

\begin{figure*}[!t]
\centering
\subfloat[DRSN~\cite{DRSN}]
	{
	\includegraphics[width=2.2in]{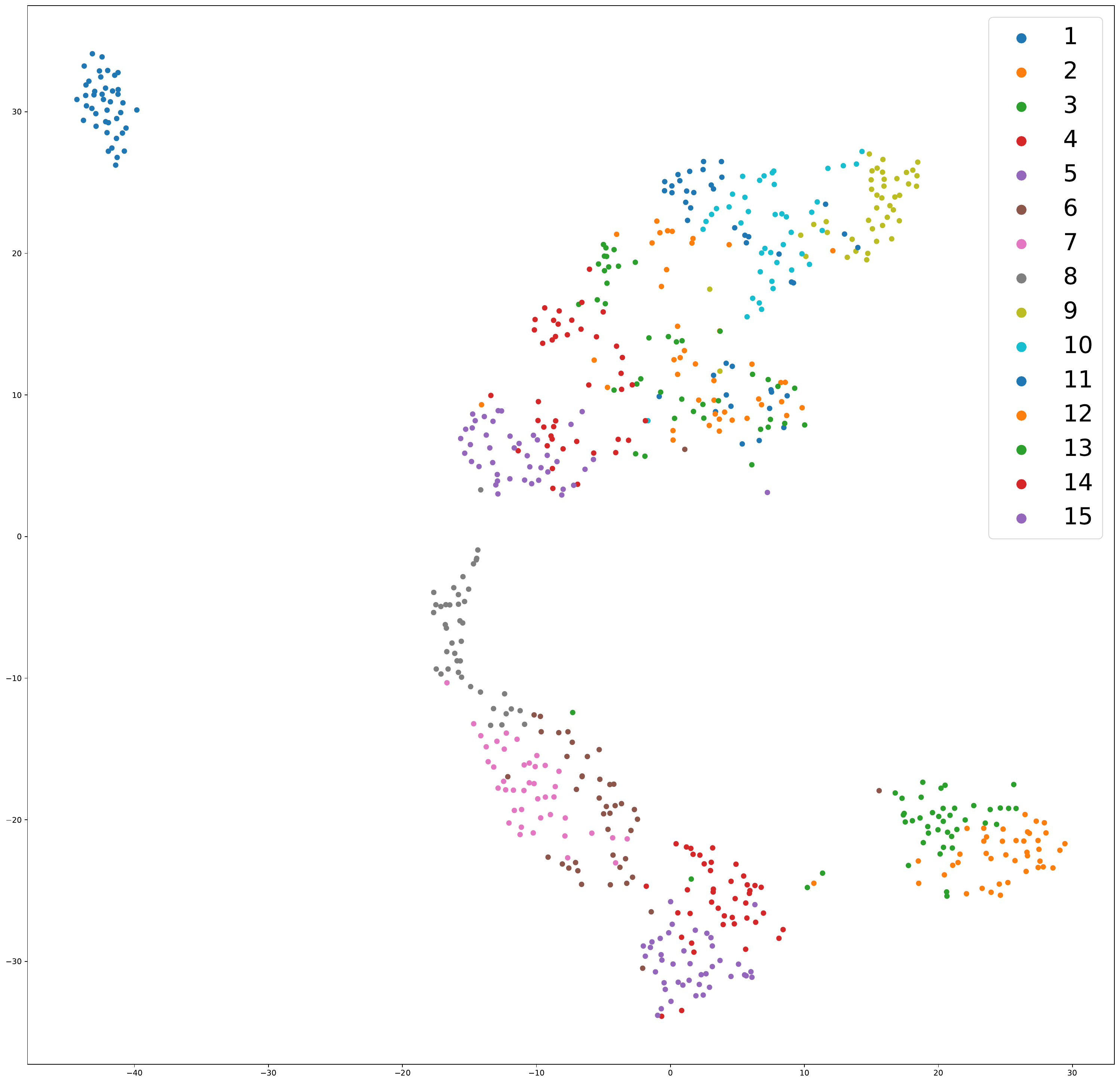}
	}
\subfloat[MLR-SNN~\cite{MLR-SNN}]
	{
	\includegraphics[width=2.2in]{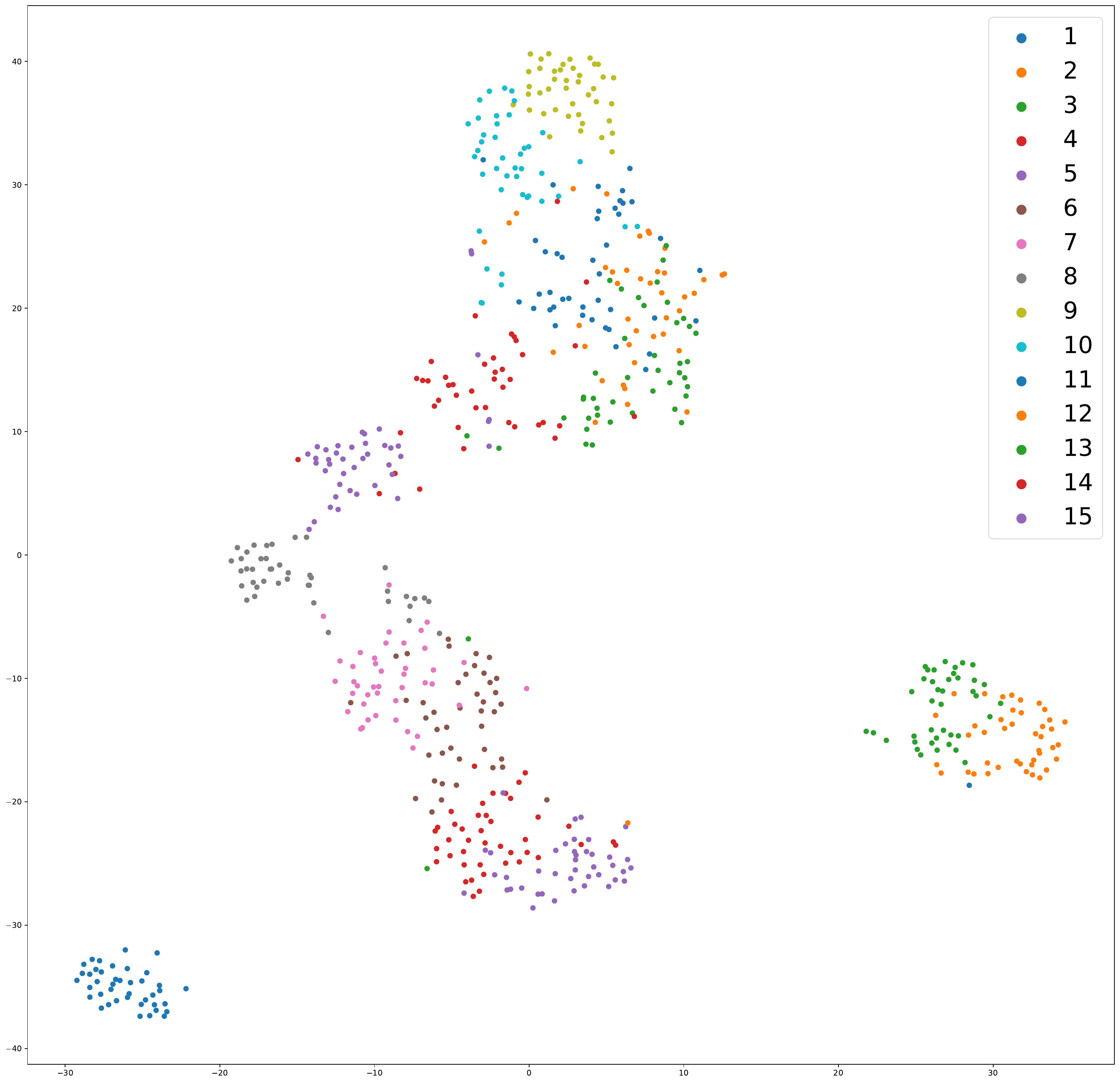}
	}
\subfloat[MRA-SNN (Ours)]
	{
	\includegraphics[width=2.2in]{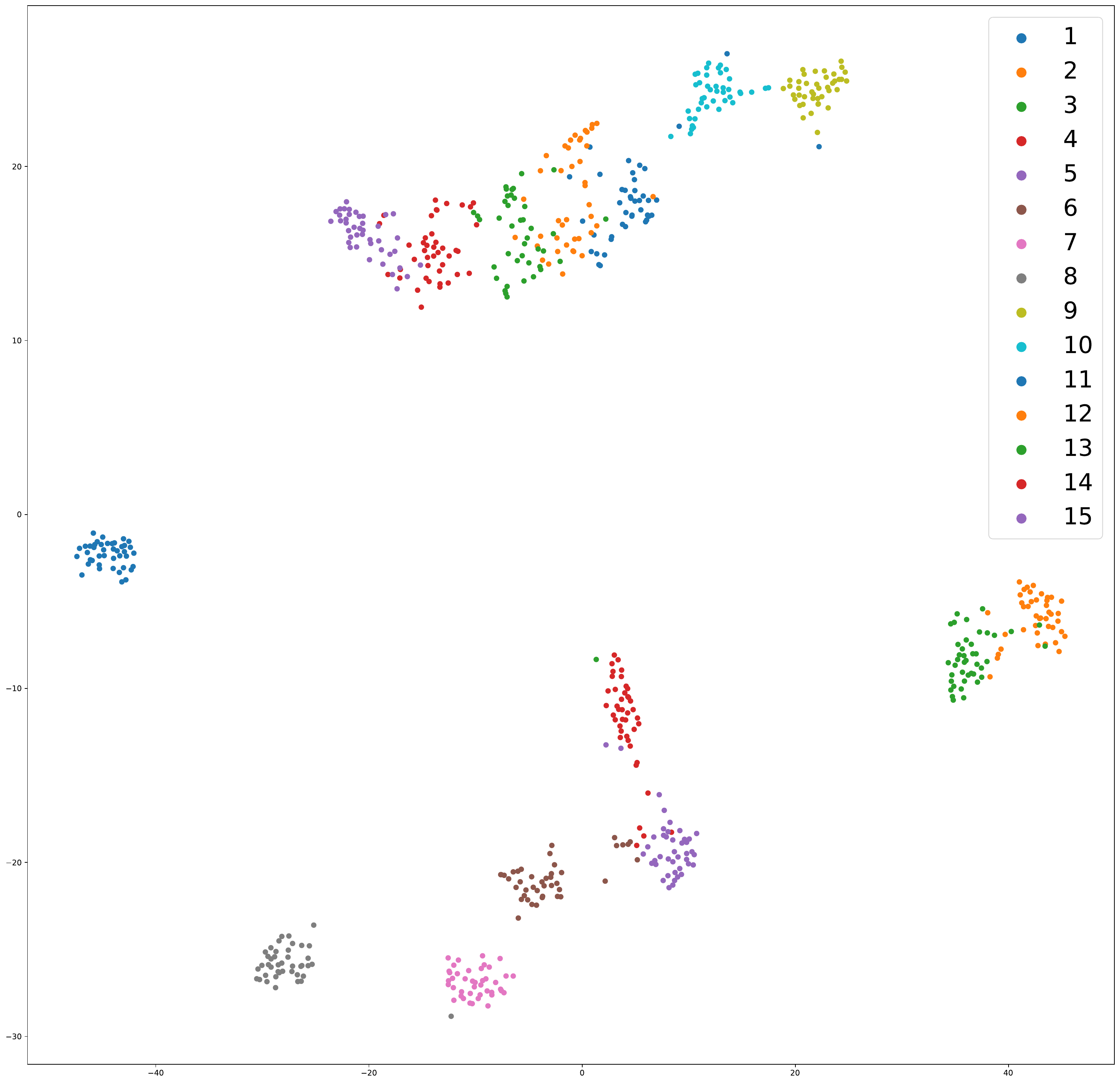}
	}
\vskip -0.1in
\caption{Two-dimensional t-SNE visualization on the MFPT dataset. The clusters of MRA-SNN are more dispersed than those of DRSN and MLR-SNN, indicating that MRA-SNN is more capable of distinguishing fault classes.}
\label{MFPtsne}
\vskip -0.12in
\end{figure*}

Comparative results with other fault diagnosis methods are shown in Table~\ref{com_result}. With only 1.75M parameters, MRA-SNN outperforms the other comparative methods on all four benchmarks. In particular, the power consumption evaluated on the MFPT benchmark is shown in Table~\ref{com_result}, where the MRA-SNN requires only 0.05G $pJ$ of energy, demonstrating remarkable energy advantages. Compared to the lightweight ANN models~\cite{LEFE-Net},~\cite{10496469}, and~\cite{LiConvFormer}, the efficient computing paradigm allows MRA-SNN to consume much less power than these methods, even though MRA-SNN has more parameters. Notably, like ANNs, MRA-SNN can be further optimized using lightweighting techniques such as quantization and distillation, making it highly exploitable.

To further demonstrate the performance of the MRA-SNN, Fig.~\ref{MFPtsne} illustrates the output 2D t-distributed stochastic neighbor embedding (t-SNE) visualization results of MRA-SNN with DRSN~\cite{DRSN} and MLR-SNN~\cite{MLR-SNN}. For DRSN and MLR-SNN, the t-SNE visualization results of most fault types are clustered together, making it difficult to distinguish the faults. In contrast, MRA-SNN is able to relatively separate fault types, and the clusters are more dispersed, indicating that it is more capable of distinguishing between fault types. The t-SNE visualization results on the JNU dataset can be found in \textbf{Appendix~\ref{AVA}}, which also shows that the MRA-SNN has a better discriminative ability.

\vspace{-0.1in}
\subsection{Ablation Study}
We conducted ablation studies to investigate the effectiveness of the components of the proposed method with comparative models as follows:

$\bullet$ w/o ASN: Vanilla LIF neurons are used in the MRA-SNN instead of the proposed attention spiking neurons.

$\bullet$ MRA-ADM: Within LIF neurons using ADM~\cite{PASNN} instead of the proposed lightweight attention mechanism.

$\bullet$ w/o RA: The spike residual attention blocks in MRA-SNN are replaced with the vanilla spike residual block.

$\bullet$ w/o CA: The channel attention mechanism in the multi-scale attention encoding module has been removed.

\begin{table}[t]
 \caption{Ablation studies of each component in the MRA-SNN\label{ADM_MFPT}}
 \vskip -0.1in
 \centering
 \begin{tabular}{cccc}
 \hline
 \multirow{2}{*}{Variant} & \multicolumn{2}{c}{Accuracy $\pm$ std (\%)} & \multirow{2}{*}{Parameters}\\
 & MFPT & JNU\\
 \hline
  w/o ASN & 91.49 $\pm$ 0.69 & 93.93 $\pm$ 0.49 & 1748319/1748304 \\
  MRA-ADM & 92.06 $\pm$ 0.91 & \textbf{96.40} $\pm$ 1.05 & 6615391/6615376 \\
  w/o RA & 93.27 $\pm$ 0.29 & 91.42 $\pm$ 1.73 & 1748385/1748370\\
  w/o CA & 92.98 $\pm$ 1.19 & 92.13 $\pm$ 1.20 & 1748402/1748387\\
  \hline
  \textbf{MRA-SNN} & \textbf{94.57} $\pm$ 0.82 & 94.18 $\pm$ 0.29 & 1748409/1748394 \\
  \hline
 \end{tabular}
\vskip -0.2in
\end{table}

The ablation results are shown in Table~\ref{ADM_MFPT}. It can be seen that using vanilla LIF neurons,  w/o ASN with only 1.75M parameters achieves an average accuracy of 91.49\% and 93.93\% on MFPT and JNU, respectively. This has exceeded the performance of the comparative methods (see Tables~\ref{com_result}), indicating that the MRA-SNN architecture is also effective for vanilla LIF neurons. ADM~\cite{PASNN} achieves better performance, but a significant increase in the number of parameters, by using heavy convolution operations as the attention mechanism. The proposed attention spiking neurons utilize lightweight channel attention and spatial attention, and the number of parameters increases by only 90 compared to  w/o ASN. This negligible parameter overhead yields performance gains of 3.08\% and 0.25\%, respectively. It is worth noting that MRA-SNN performs better than MRA-ADM on MFPT, suggesting that separate channel and spatial attention has the ability to capture more salient channel and spatial features compared to full convolution, consistent with the conclusions in~\cite{eca}.

\begin{table}[!t]
 \centering
 \caption{The influence of the order of attention\label{ablation_attention}}
 \vskip -0.1in
 \tabcolsep=0.07\columnwidth
 \begin{tabular}{ccc}
  \hline
 Order & MFPT & JNU\\
  \hline
  SA-CA & 94.61\% & 94.22\%\\
  Parallel & 94.44\% & 94.11\%\\
  CA-SA (Deafult) & 94.57\% & 94.18\%\\
  \hline
 \end{tabular}
\vskip -0.12in
\end{table}

\begin{table}[!t]
 \centering
 \caption{The influence of convolution kernel size in multi-scale coding on the MFPT dataset\label{ablation_kernel}}
 \vskip -0.1in
 \tabcolsep=0.04\columnwidth
 \begin{tabular}{cc|cc}
  \hline
 Kernel size & Acc (\%) & Kernel size & Acc (\%)\\
  \hline
  (1,3,5) & 93.65 & (1,3,9) & 94.76\\
  (1,3,7) & 93.49 & (1,5,9) & 92.86\\
  (1,5,7) & 93.65 & (3,5,9) & 94.44\\
  (3,5,7) (Deafult) & 94.57 & (5,7,9) & 95.24\\
  \hline
 \end{tabular}
\vskip -0.12in
\end{table}

\textbf{Influence of the attention structure.} In addition, we ablated the attention structure in the MRA-SNN and the resulting w/o RA and w/o CA still achieved excellent accuracy. In particular, the accuracies of w/o RA and w/o CA on the MFPT exceed the highest accuracy of the comparative models, demonstrating that our deliberately designed multi-scale attention encoding module and the spike residual attention module alone can also achieve promising performance.

\textbf{Influence of attention order and multi-scale convolution kernels.} The influence of MRA-SNN on the order of attention and the size of the multi-scale convolution kernel is examined in Table~\ref{ablation_attention} and Table~\ref{ablation_kernel}, respectively. The results show that MRA-SNN consistently delivers superior performance, indicating that the MRA-SNN architecture is insensitive to these hyperparameters and does not need to be deliberately tuned to achieve satisfactory performance.

\subsection{Robustness Evaluation and Comparison}

\begin{table}[!t]
\tabcolsep=0.01\columnwidth
 \centering
 \caption{Comparative results (\%) on MFPT under different SNRs (dB)\label{noise_MFPT}}
 \vskip -0.1in
 \begin{tabular}{ccccccc}
  \hline
 SNR & ResNet & DRSN & DSRSN & MLR-SNN & \textbf{MRA-SNN}\\
  \hline
  30 & 83.81$\pm$0.37 & 88.73$\pm$0.21 & 86.06$\pm$0.17 & 87.98$\pm$0.22 & \textbf{93.35}$\pm$0.07\\
  25 & 83.48$\pm$0.23 & 88.31$\pm$0.16 & 85.95$\pm$0.16 & 87.72$\pm$0.22 & \textbf{92.98}$\pm$0.15\\
  20 & 82.51$\pm$0.45 & 87.56$\pm$0.35 & 84.86$\pm$0.13 & 87.16$\pm$0.44 & \textbf{92.86}$\pm$0.11\\
  15 & 75.24$\pm$0.94 & 82.31$\pm$0.27 & 80.79$\pm$0.34 & 83.31$\pm$0.19 & \textbf{91.88}$\pm$0.14\\
  10 & 54.09$\pm$0.44 & 67.78$\pm$0.18 & 64.34$\pm$0.64 & 63.33$\pm$0.36 & \textbf{81.61}$\pm$0.26\\
  5 & 31.59$\pm$0.29 & 35.19$\pm$0.41 & 41.02$\pm$0.77 & 39.36$\pm$0.77 & \textbf{58.73}$\pm$0.46\\
  0 & 15.71$\pm$0.78 & 19.85$\pm$0.22 & 13.08$\pm$0.25 & 14.07$\pm$0.15 & \textbf{25.97}$\pm$0.49\\
  \hline
 \end{tabular}
\vskip -0.12in
\end{table}

\begin{table}[!t]
\tabcolsep=0.01\columnwidth
 \centering
 \caption{Comparative results (\%) on JNU under different SNRs (dB)\label{noise_JNU}}
 \vskip -0.1in
 \begin{tabular}{ccccccc}
  \hline
 SNR & ResNet & DRSN & DSRSN & MLR-SNN & \textbf{MRA-SNN}\\
  \hline
  30 & 86.63$\pm$0.21 & 92.69$\pm$0.11 & 91.86$\pm$0.09 & 85.24$\pm$0.14  & \textbf{93.67}$\pm$0.04\\
  25 & 86.02$\pm$0.11 & 91.61$\pm$0.41 & 90.89$\pm$0.08 & 84.69$\pm$0.19 & \textbf{93.48}$\pm$0.06\\
  20 & 80.67$\pm$0.23 & 83.01$\pm$0.11 & 82.75$\pm$0.26& 79.48$\pm$0.30 & \textbf{91.13}$\pm$0.09\\
  15 & 64.57$\pm$0.12 & 65.05$\pm$0.42 & 66.29$\pm$0.08& 61.07$\pm$0.21 & \textbf{71.53}$\pm$0.13\\
  10 & 50.28$\pm$0.13 & 49.82$\pm$0.21 & 54.11$\pm$0.34& 49.04$\pm$0.21 & \textbf{57.36}$\pm$0.04\\
  5 & 34.87$\pm$0.39 & 30.88$\pm$0.09 & 40.19$\pm$0.19 & 31.03$\pm$0.18 & \textbf{43.76}$\pm$0.11\\
  0 & 19.40$\pm$0.17 & 11.51$\pm$0.13 & 19.72$\pm$0.31 & 9.85$\pm$0.22 & \textbf{20.18}$\pm$0.33\\
  \hline
 \end{tabular}
\vskip -0.2in
\end{table}

Due to the harshness of the actual working conditions, the vibration signals are inevitably affected by noise. The noise robustness of fault diagnosis algorithms is extremely critical for deployment in real-world scenarios. To evaluate the noise robustness of the proposed method, we add noise to the raw vibration signal to obtain different signal-to-noise ratios (SNR). For specific details on adding noise, see \textbf{Appendix~\ref{addnoise}}. The noise robustness evaluation and comparative results on the MFPT and JNU datasets are shown in Table~\ref{noise_MFPT} and Table~\ref{noise_JNU}, respectively. When the noise influence is weak (SNR greater than 20 dB), the performance of the different fault diagnosis models is slightly affected. As the noise increases, the performance of the diagnosis models degrades dramatically, especially for ResNet on MFPT. Compared to other comparative models, the proposed MRA-SNN has consistently higher diagnostic accuracy at any SNR. Additional visualizations and accuracy change curves during noise interference are provided in \textbf{Appendix~\ref{com_cm}}, which more visually illustrates the noise robustness of the MRA-SNN. The robust diagnostic performance under noise interference demonstrates that our MRA-SNN can be better applied in real-world scenarios, opening up further opportunities for deployment.

\subsection{Output Visualization}
\begin{figure}[t]
\centering
\subfloat[Class 0 (normal)]
	{
	\includegraphics[width=1.6in]{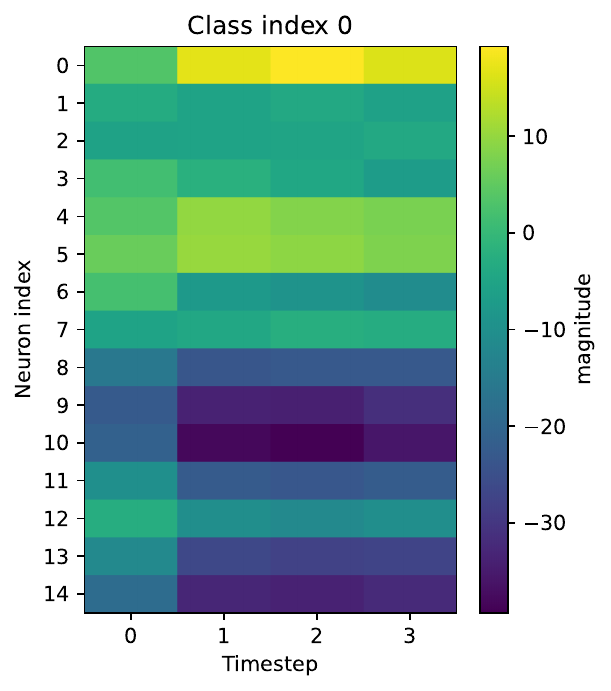}
	}
\subfloat[Class 3 (fault 2)]
	{
	\includegraphics[width=1.6in]{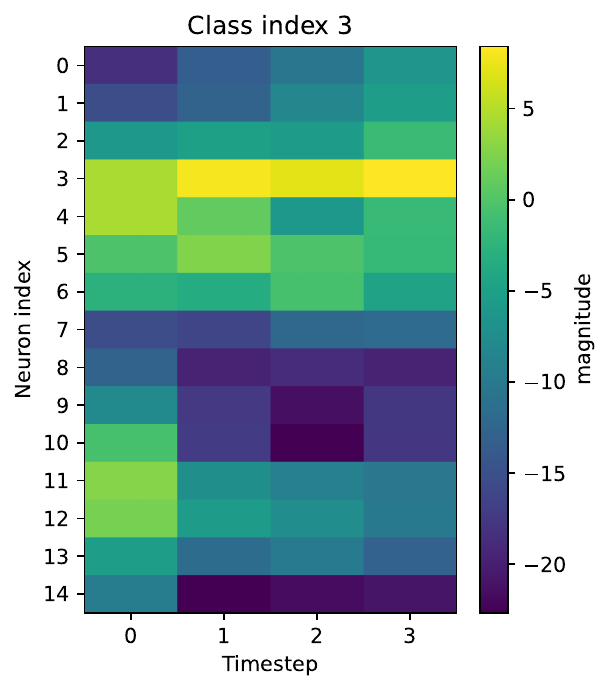}
	}

\vskip -0.15in
\caption{Visualization of MRA-SNN output in 4 timesteps. For each fault class, except for the first timestep accumulating the membrane potential, MRA-SNN generates discriminative output for the target class at subsequent timesteps.}
\label{out_vis}
\vskip -0.2in
\end{figure}
To make the biological plausibility of MRA-SNN more intuitive, we visualized its output in four timesteps, as shown in Fig.~\ref{out_vis}. For samples belonging to target classes 0 and 3, the output neurons at the corresponding positions all generated larger predictive values. For the other non-target classes, MRA-SNN generated significantly smaller predictions, which exhibited excellent distinguishability. It is worth noting that the predicted values output by MRA-SNN at the first timestep (indexed at 0) are not discriminative. The explanation for this is that the SNN is just accumulating the membrane potential at the first timestep and therefore produces few and unstable spikes. As the timestep continues, the SNN is able to make accurate predictions (the last three timesteps in Fig.~\ref{out_vis}).

\vspace{-0.05in}
\subsection{Influence of Timestep}

\begin{figure}[!t]
\centering
\includegraphics[width=2.5in]{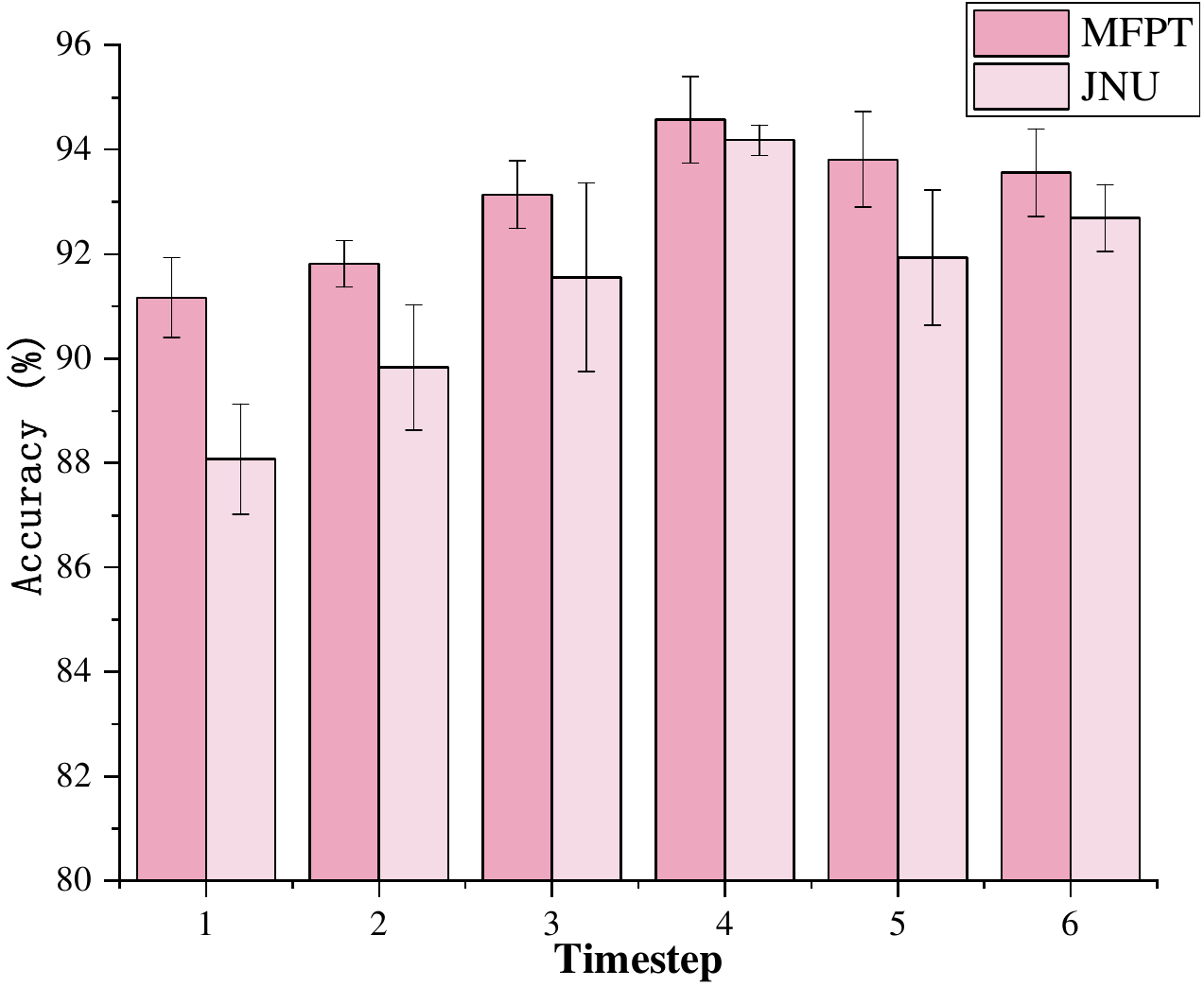}
\vskip -0.15in
\caption{Diagnostic performance of MRA-SNN with different timesteps. At ultra-low latency (timestep of 1), MRA-SNN still achieves satisfactory diagnosis performance.}
\label{timestep}
\vskip -0.2in
\end{figure}

We evaluate the performance of the MRA-SNN with different timesteps to validate its performance efficiency trade-off. The results are shown in Fig.~\ref{timestep}. As the timestep increases, the overall performance of MRA-SNN gradually improves. When the timestep is 4, the performance of MRA-SNN saturates and degrades as the timestep increases. It is worth noting that even with a timestep of 1, the average accuracy of MRA-SNN on MFPT is still above 91.17\%, which exceeds the other comparative models in Table~\ref{com_result}. In this case, the MRA-SNN provides an excellent balance of performance and efficiency. For JNU, the influence of the timestep on MRA-SNN is more pronounced, but acceptable diagnostic performance is still achieved with a timestep of 1 (which exceeds that of Spiking ResNet and MLR-SNN with a timestep of 4 in Table~\ref{com_result}).

\section{Discussion}
\label{Discussion}
\textbf{Differences from existing work.} The basic operations of both SNNs and ANNs are convolution and fully connected operators, which makes the MRA-SNN somewhat similar to ANN. Although the macro-architecture of the MRA-SNN is derived from the multi-scale and residual modules, we integrate them for the first time with SNNs in industrial scenarios to improve both performance and efficiency. In particular, when integrating these two modules into the SNN, we intentionally preserve the binary spike output, which enables deployment on a neuromorphic chip and facilitates application in real-world scenarios. In addition, existing work combines SNNs with attention at the architectural level~\cite{10032591}, while we propose attention neurons that are easier to integrate on neuromorphic chips.  Therefore, while our macro-architecture is a pre-existing style, the spike-driven modular design and our innovative attention spiking neuron allow MRA-SNN to outperform existing methods in terms of performance, efficiency, power consumption, and robustness.

\textbf{Feasible extensions.} In terms of efficiency, MRA-SNN can be further facilitated by knowledge distillation, network pruning, quantization, and other techniques; in terms of performance, although preprocessing is eliminated, combining minimal preprocessing MRA-SNN should further improve its accuracy and robustness without sacrificing efficiency. We consider these to be future studies and will release the code to facilitate further community research.

\vspace{-0.08in}
\section{Conclusion}
\label{Conclusion}
In this paper, we facilitate the application of SNNs to industrial scenarios by proposing MRA-SNN for end-to-end bearing fault diagnosis. MRA-SNN offers a lightweight architecture, superior spike encoding and feature extraction capabilities for efficient and effective fault diagnosis without pre-processing vibration signals. In addition, a lightweight attention spiking neuron that mimics biological synaptic filtering through a separated channel-spatial attention mechanism and enhances the performance and robustness of the MRA-SNN was presented. Extensive experiments on four benchmarks show that MRA-SNN outperforms existing SNN fault diagnosis methods, and even ANN models, in terms of performance and noise robustness. We expect that this will facilitate the application and deployment of SNNs in more real-world scenarios.

\section*{Acknowledgments}

This work was supported by the National Natural Science Foundation of China (Grants 62276054 and 62406060) and the Sichuan Science and Technology Program (Grant 2025ZNSFSC1500).

\bibliographystyle{ACM-Reference-Format}
\bibliography{main}


\begin{thebibliography}{47}


\ifx \showCODEN    \undefined \def \showCODEN     #1{\unskip}     \fi
\ifx \showDOI      \undefined \def \showDOI       #1{#1}\fi
\ifx \showISBNx    \undefined \def \showISBNx     #1{\unskip}     \fi
\ifx \showISBNxiii \undefined \def \showISBNxiii  #1{\unskip}     \fi
\ifx \showISSN     \undefined \def \showISSN      #1{\unskip}     \fi
\ifx \showLCCN     \undefined \def \showLCCN      #1{\unskip}     \fi
\ifx \shownote     \undefined \def \shownote      #1{#1}          \fi
\ifx \showarticletitle \undefined \def \showarticletitle #1{#1}   \fi
\ifx \showURL      \undefined \def \showURL       {\relax}        \fi
\providecommand\bibfield[2]{#2}
\providecommand\bibinfo[2]{#2}
\providecommand\natexlab[1]{#1}
\providecommand\showeprint[2][]{arXiv:#2}

\bibitem[Bal and Sengupta(2024)]%
        {bal2024spikingbert}
\bibfield{author}{\bibinfo{person}{Malyaban Bal} {and}
  \bibinfo{person}{Abhronil Sengupta}.} \bibinfo{year}{2024}\natexlab{}.
\newblock \showarticletitle{Spikingbert: Distilling bert to train spiking
  language models using implicit differentiation}. In
  \bibinfo{booktitle}{\emph{Proceedings of the AAAI conference on artificial
  intelligence}}, Vol.~\bibinfo{volume}{38}. \bibinfo{pages}{10998--11006}.
\newblock


\bibitem[Chen et~al\mbox{.}(2021)]%
        {chen2021bearing}
\bibfield{author}{\bibinfo{person}{Xiaohan Chen}, \bibinfo{person}{Beike
  Zhang}, {and} \bibinfo{person}{Dong Gao}.} \bibinfo{year}{2021}\natexlab{}.
\newblock \showarticletitle{Bearing fault diagnosis base on multi-scale CNN and
  LSTM model}.
\newblock \bibinfo{journal}{\emph{J. Intell. Manuf.}}  \bibinfo{volume}{32}
  (\bibinfo{year}{2021}), \bibinfo{pages}{971--987}.
\newblock


\bibitem[Ding et~al\mbox{.}(2024)]%
        {SSNN}
\bibfield{author}{\bibinfo{person}{Yongqi Ding}, \bibinfo{person}{Lin Zuo},
  \bibinfo{person}{Mengmeng Jing}, \bibinfo{person}{Pei He}, {and}
  \bibinfo{person}{Yongjun Xiao}.} \bibinfo{year}{2024}\natexlab{}.
\newblock \showarticletitle{Shrinking Your TimeStep: Towards Low-Latency
  Neuromorphic Object Recognition with Spiking Neural Networks}. In
  \bibinfo{booktitle}{\emph{Proceedings of the {AAAI} {Conference} on
  {Artificial} {Intelligence}}}. \bibinfo{pages}{11811--11819}.
\newblock


\bibitem[Ding et~al\mbox{.}(2023)]%
        {PASNN}
\bibfield{author}{\bibinfo{person}{Yongqi Ding}, \bibinfo{person}{Lin Zuo},
  \bibinfo{person}{Kunshan Yang}, \bibinfo{person}{Zhongshu Chen},
  \bibinfo{person}{Jian Hu}, {and} \bibinfo{person}{Tangfan Xiahou}.}
  \bibinfo{year}{2023}\natexlab{}.
\newblock \showarticletitle{An improved probabilistic spiking neural network
  with enhanced discriminative ability}.
\newblock \bibinfo{journal}{\emph{Knowledge-Based Syst.}}
  \bibinfo{volume}{280} (\bibinfo{year}{2023}), \bibinfo{pages}{111024}.
\newblock


\bibitem[Fang et~al\mbox{.}(2021a)]%
        {LEFE-Net}
\bibfield{author}{\bibinfo{person}{Hairui Fang}, \bibinfo{person}{Jin Deng},
  \bibinfo{person}{Bo Zhao}, \bibinfo{person}{Yan Shi}, \bibinfo{person}{Jianye
  Zhou}, {and} \bibinfo{person}{Siyu Shao}.} \bibinfo{year}{2021}\natexlab{a}.
\newblock \showarticletitle{LEFE-Net: A Lightweight Efficient Feature
  Extraction Network With Strong Robustness for Bearing Fault Diagnosis}.
\newblock \bibinfo{journal}{\emph{IEEE Transactions on Instrumentation and
  Measurement}}  \bibinfo{volume}{70} (\bibinfo{year}{2021}),
  \bibinfo{pages}{1--11}.
\newblock


\bibitem[Fang et~al\mbox{.}(2021b)]%
        {PLIF}
\bibfield{author}{\bibinfo{person}{Wei Fang}, \bibinfo{person}{Zhaofei Yu},
  \bibinfo{person}{Yanqi Chen}, \bibinfo{person}{Timoth\'ee Masquelier},
  \bibinfo{person}{Tiejun Huang}, {and} \bibinfo{person}{Yonghong Tian}.}
  \bibinfo{year}{2021}\natexlab{b}.
\newblock \showarticletitle{Incorporating Learnable Membrane Time Constant To
  Enhance Learning of Spiking Neural Networks}. In
  \bibinfo{booktitle}{\emph{Proc. IEEE/CVF Int. Conf. Comput. Vis. (ICCV)}}.
  \bibinfo{pages}{2661--2671}.
\newblock


\bibitem[Feldmann et~al\mbox{.}(2019)]%
        {feldmann2019all}
\bibfield{author}{\bibinfo{person}{Johannes Feldmann}, \bibinfo{person}{Nathan
  Youngblood}, \bibinfo{person}{C~David Wright}, \bibinfo{person}{Harish
  Bhaskaran}, {and} \bibinfo{person}{Wolfram~HP Pernice}.}
  \bibinfo{year}{2019}\natexlab{}.
\newblock \showarticletitle{All-optical spiking neurosynaptic networks with
  self-learning capabilities}.
\newblock \bibinfo{journal}{\emph{Nature}} \bibinfo{volume}{569},
  \bibinfo{number}{7755} (\bibinfo{year}{2019}), \bibinfo{pages}{208--214}.
\newblock


\bibitem[He et~al\mbox{.}(2016)]%
        {ResNet}
\bibfield{author}{\bibinfo{person}{Kaiming He}, \bibinfo{person}{Xiangyu
  Zhang}, \bibinfo{person}{Shaoqing Ren}, {and} \bibinfo{person}{Jian Sun}.}
  \bibinfo{year}{2016}\natexlab{}.
\newblock \showarticletitle{Deep Residual Learning for Image Recognition}. In
  \bibinfo{booktitle}{\emph{Proc. IEEE/CVF Conf. Comput. Vis. Pattern Recognit.
  (CVPR)}}.
\newblock


\bibitem[Horowitz(2014)]%
        {6757323}
\bibfield{author}{\bibinfo{person}{Mark Horowitz}.}
  \bibinfo{year}{2014}\natexlab{}.
\newblock \showarticletitle{1.1 Computing's energy problem (and what we can do
  about it)}. In \bibinfo{booktitle}{\emph{2014 IEEE International Solid-State
  Circuits Conference Digest of Technical Papers (ISSCC)}}.
  \bibinfo{pages}{10--14}.
\newblock


\bibitem[Hu et~al\mbox{.}(2018)]%
        {SE}
\bibfield{author}{\bibinfo{person}{Jie Hu}, \bibinfo{person}{Li Shen}, {and}
  \bibinfo{person}{Gang Sun}.} \bibinfo{year}{2018}\natexlab{}.
\newblock \showarticletitle{Squeeze-and-Excitation Networks}. In
  \bibinfo{booktitle}{\emph{Proceedings of the IEEE Conference on Computer
  Vision and Pattern Recognition (CVPR)}}.
\newblock


\bibitem[Hu et~al\mbox{.}(2022)]%
        {9900453}
\bibfield{author}{\bibinfo{person}{S.~G. Hu}, \bibinfo{person}{G.~C. Qiao},
  \bibinfo{person}{X.~K. Liu}, \bibinfo{person}{Y.~H. Liu},
  \bibinfo{person}{C.~M. Zhang}, \bibinfo{person}{Yue Zuo},
  \bibinfo{person}{Pujun Zhou}, \bibinfo{person}{Y.~A. Liu},
  \bibinfo{person}{Ning Ning}, \bibinfo{person}{Qi Yu}, {and}
  \bibinfo{person}{Yang Liu}.} \bibinfo{year}{2022}\natexlab{}.
\newblock \showarticletitle{A Co-Designed Neuromorphic Chip With Compact (17.9K
  F2) and Weak Neuron Number-Dependent Neuron/Synapse Modules}.
\newblock \bibinfo{journal}{\emph{IEEE Transactions on Biomedical Circuits and
  Systems}} \bibinfo{volume}{16}, \bibinfo{number}{6} (\bibinfo{year}{2022}),
  \bibinfo{pages}{1250--1260}.
\newblock


\bibitem[Hu et~al\mbox{.}(2024)]%
        {MSResNet}
\bibfield{author}{\bibinfo{person}{Yifan Hu}, \bibinfo{person}{Lei Deng},
  \bibinfo{person}{Yujie Wu}, \bibinfo{person}{Man Yao}, {and}
  \bibinfo{person}{Guoqi Li}.} \bibinfo{year}{2024}\natexlab{}.
\newblock \showarticletitle{Advancing Spiking Neural Networks Toward Deep
  Residual Learning}.
\newblock \bibinfo{journal}{\emph{IEEE Transactions on Neural Networks and
  Learning Systems}} (\bibinfo{year}{2024}), \bibinfo{pages}{1--15}.
\newblock


\bibitem[Ioffe and Szegedy(2015)]%
        {BN}
\bibfield{author}{\bibinfo{person}{Sergey Ioffe} {and}
  \bibinfo{person}{Christian Szegedy}.} \bibinfo{year}{2015}\natexlab{}.
\newblock \showarticletitle{Batch Normalization: Accelerating Deep Network
  Training by Reducing Internal Covariate Shift}. In
  \bibinfo{booktitle}{\emph{Proc. Int. Conf. Mach. Learn. (ICML)}},
  Vol.~\bibinfo{volume}{37}. \bibinfo{pages}{448--456}.
\newblock


\bibitem[Kim et~al\mbox{.}(2020)]%
        {Spiking_YOLO}
\bibfield{author}{\bibinfo{person}{Seijoon Kim}, \bibinfo{person}{Seongsik
  Park}, \bibinfo{person}{Byunggook Na}, {and} \bibinfo{person}{Sungroh Yoon}.}
  \bibinfo{year}{2020}\natexlab{}.
\newblock \showarticletitle{Spiking-yolo: spiking neural network for
  energy-efficient object detection}. In \bibinfo{booktitle}{\emph{Proc. AAAI
  Conf. Artif. Intell.}}, Vol.~\bibinfo{volume}{34}.
  \bibinfo{pages}{11270--11277}.
\newblock


\bibitem[Kundu et~al\mbox{.}(2021)]%
        {Kundu_2021_WACV}
\bibfield{author}{\bibinfo{person}{Souvik Kundu}, \bibinfo{person}{Gourav
  Datta}, \bibinfo{person}{Massoud Pedram}, {and} \bibinfo{person}{Peter~A.
  Beerel}.} \bibinfo{year}{2021}\natexlab{}.
\newblock \showarticletitle{Spike-Thrift: Towards Energy-Efficient Deep Spiking
  Neural Networks by Limiting Spiking Activity via Attention-Guided
  Compression}. In \bibinfo{booktitle}{\emph{Proceedings of the IEEE/CVF Winter
  Conference on Applications of Computer Vision (WACV)}}.
  \bibinfo{pages}{3953--3962}.
\newblock


\bibitem[Li et~al\mbox{.}(2013)]%
        {JNU}
\bibfield{author}{\bibinfo{person}{Ke Li}, \bibinfo{person}{Xueliang Ping},
  \bibinfo{person}{Huaqing Wang}, \bibinfo{person}{Peng Chen}, {and}
  \bibinfo{person}{Yi Cao}.} \bibinfo{year}{2013}\natexlab{}.
\newblock \showarticletitle{Sequential Fuzzy Diagnosis Method for Motor Roller
  Bearing in Variable Operating Conditions Based on Vibration Analysis}.
\newblock \bibinfo{journal}{\emph{Sensors}} \bibinfo{volume}{13},
  \bibinfo{number}{6} (\bibinfo{year}{2013}), \bibinfo{pages}{8013--8041}.
\newblock


\bibitem[Li et~al\mbox{.}(2021)]%
        {li2021intelligent}
\bibfield{author}{\bibinfo{person}{Yongbo Li}, \bibinfo{person}{Shun Wang},
  {and} \bibinfo{person}{Zichen Deng}.} \bibinfo{year}{2021}\natexlab{}.
\newblock \showarticletitle{Intelligent fault identification of rotary
  machinery using refined composite multi-scale Lempel--Ziv complexity}.
\newblock \bibinfo{journal}{\emph{Journal of Manufacturing Systems}}
  \bibinfo{volume}{61} (\bibinfo{year}{2021}), \bibinfo{pages}{725--735}.
\newblock


\bibitem[Luo et~al\mbox{.}(2021)]%
        {luo2021siamsnn}
\bibfield{author}{\bibinfo{person}{Yihao Luo}, \bibinfo{person}{Min Xu},
  \bibinfo{person}{Caihong Yuan}, \bibinfo{person}{Xiang Cao},
  \bibinfo{person}{Liangqi Zhang}, \bibinfo{person}{Yan Xu},
  \bibinfo{person}{Tianjiang Wang}, {and} \bibinfo{person}{Qi Feng}.}
  \bibinfo{year}{2021}\natexlab{}.
\newblock \showarticletitle{Siamsnn: Siamese spiking neural networks for
  energy-efficient object tracking}. In \bibinfo{booktitle}{\emph{International
  Conference on Artificial Neural Networks}}. Springer,
  \bibinfo{pages}{182--194}.
\newblock


\bibitem[Magee(2000)]%
        {Magee2000DendriticIO}
\bibfield{author}{\bibinfo{person}{J. Magee}.} \bibinfo{year}{2000}\natexlab{}.
\newblock \showarticletitle{Dendritic integration of excitatory synaptic
  input}.
\newblock \bibinfo{journal}{\emph{Nat. Rev. Neurosci.}}  \bibinfo{volume}{1}
  (\bibinfo{year}{2000}), \bibinfo{pages}{181--190}.
\newblock


\bibitem[MFPT(2024)]%
        {MFPT}
\bibfield{author}{\bibinfo{person}{MFPT}.} \bibinfo{year}{2024}\natexlab{}.
\newblock \bibinfo{title}{Failure prevention technology website}.
\newblock \bibinfo{howpublished}{\url{https://www.mfpt.org/fault-data-sets/}}.
\newblock
\newblock
\shownote{Accessed 2024}.


\bibitem[Molchanov et~al\mbox{.}(2017)]%
        {molchanov2017pruning}
\bibfield{author}{\bibinfo{person}{Pavlo Molchanov}, \bibinfo{person}{Stephen
  Tyree}, \bibinfo{person}{Tero Karras}, \bibinfo{person}{Timo Aila}, {and}
  \bibinfo{person}{Jan Kautz}.} \bibinfo{year}{2017}\natexlab{}.
\newblock \showarticletitle{Pruning Convolutional Neural Networks for Resource
  Efficient Inference}. In \bibinfo{booktitle}{\emph{International Conference
  on Learning Representations}}.
\newblock


\bibitem[Qin et~al\mbox{.}(2023)]%
        {qin2023low}
\bibfield{author}{\bibinfo{person}{Lang Qin}, \bibinfo{person}{Rui Yan}, {and}
  \bibinfo{person}{Huajin Tang}.} \bibinfo{year}{2023}\natexlab{}.
\newblock \showarticletitle{A low latency adaptive coding spike framework for
  deep reinforcement learning}. In \bibinfo{booktitle}{\emph{Proceedings of the
  Thirty-Second International Joint Conference on Artificial Intelligence}}.
  \bibinfo{pages}{3049--3057}.
\newblock


\bibitem[Roy et~al\mbox{.}(2019)]%
        {roy2019towards}
\bibfield{author}{\bibinfo{person}{Kaushik Roy}, \bibinfo{person}{Akhilesh
  Jaiswal}, {and} \bibinfo{person}{Priyadarshini Panda}.}
  \bibinfo{year}{2019}\natexlab{}.
\newblock \showarticletitle{Towards spike-based machine intelligence with
  neuromorphic computing}.
\newblock \bibinfo{journal}{\emph{Nature}} \bibinfo{volume}{575},
  \bibinfo{number}{7784} (\bibinfo{year}{2019}), \bibinfo{pages}{607--617}.
\newblock


\bibitem[Serre et~al\mbox{.}(2007)]%
        {4069258}
\bibfield{author}{\bibinfo{person}{Thomas Serre}, \bibinfo{person}{Lior Wolf},
  \bibinfo{person}{Stanley Bileschi}, \bibinfo{person}{Maximilian Riesenhuber},
  {and} \bibinfo{person}{Tomaso Poggio}.} \bibinfo{year}{2007}\natexlab{}.
\newblock \showarticletitle{Robust Object Recognition with Cortex-Like
  Mechanisms}.
\newblock \bibinfo{journal}{\emph{IEEE Transactions on Pattern Analysis and
  Machine Intelligence}} \bibinfo{volume}{29}, \bibinfo{number}{3}
  (\bibinfo{year}{2007}), \bibinfo{pages}{411--426}.
\newblock


\bibitem[Shao et~al\mbox{.}(2019)]%
        {Gearbox}
\bibfield{author}{\bibinfo{person}{Siyu Shao}, \bibinfo{person}{Stephen
  McAleer}, \bibinfo{person}{Ruqiang Yan}, {and} \bibinfo{person}{Pierre
  Baldi}.} \bibinfo{year}{2019}\natexlab{}.
\newblock \showarticletitle{Highly Accurate Machine Fault Diagnosis Using Deep
  Transfer Learning}.
\newblock \bibinfo{journal}{\emph{IEEE Transactions on Industrial Informatics}}
  \bibinfo{volume}{15}, \bibinfo{number}{4} (\bibinfo{year}{2019}),
  \bibinfo{pages}{2446--2455}.
\newblock


\bibitem[Spruston et~al\mbox{.}(1994)]%
        {SPRUSTON1994161}
\bibfield{author}{\bibinfo{person}{Nelson Spruston}, \bibinfo{person}{David~B.
  Jaffe}, {and} \bibinfo{person}{Daniel Johnston}.}
  \bibinfo{year}{1994}\natexlab{}.
\newblock \showarticletitle{Dendritic attenuation of synaptic potentials and
  currents: the role of passive membrane properties}.
\newblock \bibinfo{journal}{\emph{Trends Neurosci.}} \bibinfo{volume}{17},
  \bibinfo{number}{4} (\bibinfo{year}{1994}), \bibinfo{pages}{161--166}.
\newblock


\bibitem[Su et~al\mbox{.}(2023)]%
        {Su_2023_ICCV}
\bibfield{author}{\bibinfo{person}{Qiaoyi Su}, \bibinfo{person}{Yuhong Chou},
  \bibinfo{person}{Yifan Hu}, \bibinfo{person}{Jianing Li},
  \bibinfo{person}{Shijie Mei}, \bibinfo{person}{Ziyang Zhang}, {and}
  \bibinfo{person}{Guoqi Li}.} \bibinfo{year}{2023}\natexlab{}.
\newblock \showarticletitle{Deep Directly-Trained Spiking Neural Networks for
  Object Detection}. In \bibinfo{booktitle}{\emph{Proceedings of the IEEE/CVF
  International Conference on Computer Vision (ICCV)}}.
  \bibinfo{pages}{6555--6565}.
\newblock


\bibitem[Tang et~al\mbox{.}(2021)]%
        {tang2021deep}
\bibfield{author}{\bibinfo{person}{Guangzhi Tang}, \bibinfo{person}{Neelesh
  Kumar}, \bibinfo{person}{Raymond Yoo}, {and} \bibinfo{person}{Konstantinos
  Michmizos}.} \bibinfo{year}{2021}\natexlab{}.
\newblock \showarticletitle{Deep reinforcement learning with population-coded
  spiking neural network for continuous control}. In
  \bibinfo{booktitle}{\emph{Conference on Robot Learning}}. PMLR,
  \bibinfo{pages}{2016--2029}.
\newblock


\bibitem[Wang and Li(2023)]%
        {MLR-SNN}
\bibfield{author}{\bibinfo{person}{Huan Wang} {and} \bibinfo{person}{Yan-Fu
  Li}.} \bibinfo{year}{2023}\natexlab{}.
\newblock \showarticletitle{Bioinspired membrane learnable spiking neural
  network for autonomous vehicle sensors fault diagnosis under open
  environments}.
\newblock \bibinfo{journal}{\emph{Reliab. Eng. Syst. Saf.}}
  \bibinfo{volume}{233} (\bibinfo{year}{2023}), \bibinfo{pages}{109102}.
\newblock


\bibitem[Wang et~al\mbox{.}(2022)]%
        {ISNN}
\bibfield{author}{\bibinfo{person}{Jun Wang}, \bibinfo{person}{Tianfu Li},
  \bibinfo{person}{Chuang Sun}, \bibinfo{person}{Ruqiang Yan}, {and}
  \bibinfo{person}{Xuefeng Chen}.} \bibinfo{year}{2022}\natexlab{}.
\newblock \showarticletitle{Improved spiking neural network for intershaft
  bearing fault diagnosis}.
\newblock \bibinfo{journal}{\emph{J. Manuf. Syst.}}  \bibinfo{volume}{65}
  (\bibinfo{year}{2022}), \bibinfo{pages}{208--219}.
\newblock


\bibitem[Wang et~al\mbox{.}(2023)]%
        {wang2023bearing}
\bibfield{author}{\bibinfo{person}{Pengcheng Wang}, \bibinfo{person}{Hui
  Xiong}, {and} \bibinfo{person}{Haoxiang He}.}
  \bibinfo{year}{2023}\natexlab{}.
\newblock \showarticletitle{Bearing fault diagnosis under various conditions
  using an incremental learning-based multi-task shared classifier}.
\newblock \bibinfo{journal}{\emph{Knowledge-based systems}}
  \bibinfo{volume}{266} (\bibinfo{year}{2023}), \bibinfo{pages}{110395}.
\newblock


\bibitem[Wang et~al\mbox{.}(2020)]%
        {eca}
\bibfield{author}{\bibinfo{person}{Qilong Wang}, \bibinfo{person}{Banggu Wu},
  \bibinfo{person}{Pengfei Zhu}, \bibinfo{person}{Peihua Li},
  \bibinfo{person}{Wangmeng Zuo}, {and} \bibinfo{person}{Qinghua Hu}.}
  \bibinfo{year}{2020}\natexlab{}.
\newblock \showarticletitle{ECA-Net: Efficient Channel Attention for Deep
  Convolutional Neural Networks}. In \bibinfo{booktitle}{\emph{Proc. IEEE/CVF
  Conf. Comput. Vis. Pattern Recognit. (CVPR)}}.
\newblock


\bibitem[Wang et~al\mbox{.}(2024)]%
        {10496469}
\bibfield{author}{\bibinfo{person}{Yanzhi Wang}, \bibinfo{person}{Ziyang Yu},
  \bibinfo{person}{Jinhong Wu}, \bibinfo{person}{Chu Wang}, \bibinfo{person}{Qi
  Zhou}, {and} \bibinfo{person}{Jiexiang Hu}.} \bibinfo{year}{2024}\natexlab{}.
\newblock \showarticletitle{Adaptive Knowledge Distillation-Based Lightweight
  Intelligent Fault Diagnosis Framework in IoT Edge Computing}.
\newblock \bibinfo{journal}{\emph{IEEE Internet of Things Journal}}
  \bibinfo{volume}{11}, \bibinfo{number}{13} (\bibinfo{year}{2024}),
  \bibinfo{pages}{23156--23169}.
\newblock


\bibitem[Woo et~al\mbox{.}(2018)]%
        {CBAM}
\bibfield{author}{\bibinfo{person}{Sanghyun Woo}, \bibinfo{person}{Jongchan
  Park}, \bibinfo{person}{Joon-Young Lee}, {and} \bibinfo{person}{In~So
  Kweon}.} \bibinfo{year}{2018}\natexlab{}.
\newblock \showarticletitle{CBAM: Convolutional Block Attention Module}. In
  \bibinfo{booktitle}{\emph{Proceedings of the European Conference on Computer
  Vision (ECCV)}}.
\newblock


\bibitem[Wu et~al\mbox{.}(2022)]%
        {9543525}
\bibfield{author}{\bibinfo{person}{Jibin Wu}, \bibinfo{person}{Chenglin Xu},
  \bibinfo{person}{Xiao Han}, \bibinfo{person}{Daquan Zhou},
  \bibinfo{person}{Malu Zhang}, \bibinfo{person}{Haizhou Li}, {et~al\mbox{.}}}
  \bibinfo{year}{2022}\natexlab{}.
\newblock \showarticletitle{Progressive Tandem Learning for Pattern Recognition
  With Deep Spiking Neural Networks}.
\newblock \bibinfo{journal}{\emph{{IEEE} Trans. Pattern Anal. Mach. Intell.}}
  \bibinfo{volume}{44}, \bibinfo{number}{11} (\bibinfo{year}{2022}),
  \bibinfo{pages}{7824--7840}.
\newblock


\bibitem[Wu et~al\mbox{.}(2018)]%
        {STBP}
\bibfield{author}{\bibinfo{person}{Yujie Wu}, \bibinfo{person}{Lei Deng},
  \bibinfo{person}{Guoqi Li}, \bibinfo{person}{Jun Zhu}, {and}
  \bibinfo{person}{Luping Shi}.} \bibinfo{year}{2018}\natexlab{}.
\newblock \showarticletitle{Spatio-Temporal Backpropagation for Training
  High-Performance Spiking Neural Networks}.
\newblock \bibinfo{journal}{\emph{Front. Neurosci.}}  \bibinfo{volume}{12}
  (\bibinfo{year}{2018}).
\newblock


\bibitem[Xu and Ji(2024)]%
        {10555174}
\bibfield{author}{\bibinfo{person}{Lie Xu} {and} \bibinfo{person}{Daxiong Ji}.}
  \bibinfo{year}{2024}\natexlab{}.
\newblock \showarticletitle{Online Fault Diagnosis Using Bioinspired Spike
  Neural Network}.
\newblock \bibinfo{journal}{\emph{IEEE Transactions on Industrial Informatics}}
  (\bibinfo{year}{2024}), \bibinfo{pages}{1--9}.
\newblock


\bibitem[Xu et~al\mbox{.}(2024)]%
        {DSRSN}
\bibfield{author}{\bibinfo{person}{Zongtang Xu}, \bibinfo{person}{Yumei Ma},
  \bibinfo{person}{Zhenkuan Pan}, {and} \bibinfo{person}{Xiaoyang Zheng}.}
  \bibinfo{year}{2024}\natexlab{}.
\newblock \showarticletitle{Deep Spiking Residual Shrinkage Network for Bearing
  Fault Diagnosis}.
\newblock \bibinfo{journal}{\emph{IEEE Transactions on Cybernetics}}
  \bibinfo{volume}{54}, \bibinfo{number}{3} (\bibinfo{year}{2024}),
  \bibinfo{pages}{1608--1613}.
\newblock


\bibitem[Yan et~al\mbox{.}(2024)]%
        {LiConvFormer}
\bibfield{author}{\bibinfo{person}{Shen Yan}, \bibinfo{person}{Haidong Shao},
  \bibinfo{person}{Jie Wang}, \bibinfo{person}{Xinyu Zheng}, {and}
  \bibinfo{person}{Bin Liu}.} \bibinfo{year}{2024}\natexlab{}.
\newblock \showarticletitle{LiConvFormer: A lightweight fault diagnosis
  framework using separable multiscale convolution and broadcast
  self-attention}.
\newblock \bibinfo{journal}{\emph{Expert Systems with Applications}}
  \bibinfo{volume}{237} (\bibinfo{year}{2024}), \bibinfo{pages}{121338}.
\newblock
\showISSN{0957-4174}


\bibitem[Yao et~al\mbox{.}(2024)]%
        {yao2024spike}
\bibfield{author}{\bibinfo{person}{Man Yao}, \bibinfo{person}{Ole Richter},
  \bibinfo{person}{Guangshe Zhao}, \bibinfo{person}{Ning Qiao},
  \bibinfo{person}{Yannan Xing}, \bibinfo{person}{Dingheng Wang},
  \bibinfo{person}{Tianxiang Hu}, \bibinfo{person}{Wei Fang},
  \bibinfo{person}{Tugba Demirci}, \bibinfo{person}{Michele De~Marchi},
  {et~al\mbox{.}}} \bibinfo{year}{2024}\natexlab{}.
\newblock \showarticletitle{Spike-based dynamic computing with asynchronous
  sensing-computing neuromorphic chip}.
\newblock \bibinfo{journal}{\emph{Nature Communications}} \bibinfo{volume}{15},
  \bibinfo{number}{1} (\bibinfo{year}{2024}), \bibinfo{pages}{4464}.
\newblock


\bibitem[Yao et~al\mbox{.}(2023)]%
        {10032591}
\bibfield{author}{\bibinfo{person}{Man Yao}, \bibinfo{person}{Guangshe Zhao},
  \bibinfo{person}{Hengyu Zhang}, \bibinfo{person}{Yifan Hu},
  \bibinfo{person}{Lei Deng}, \bibinfo{person}{Yonghong Tian}, {et~al\mbox{.}}}
  \bibinfo{year}{2023}\natexlab{}.
\newblock \showarticletitle{Attention Spiking Neural Networks}.
\newblock \bibinfo{journal}{\emph{{IEEE} Trans. Pattern Anal. Mach. Intell.}}
  \bibinfo{volume}{45}, \bibinfo{number}{8} (\bibinfo{year}{2023}),
  \bibinfo{pages}{9393--9410}.
\newblock


\bibitem[Yao et~al\mbox{.}(2022)]%
        {GLIF}
\bibfield{author}{\bibinfo{person}{Xingting Yao}, \bibinfo{person}{Fanrong Li},
  \bibinfo{person}{Zitao Mo}, {and} \bibinfo{person}{Jian Cheng}.}
  \bibinfo{year}{2022}\natexlab{}.
\newblock \showarticletitle{GLIF: A Unified Gated Leaky Integrate-and-Fire
  Neuron for Spiking Neural Networks}. In \bibinfo{booktitle}{\emph{Proc. Adv.
  Neural Inf. Process. Syst.}}, Vol.~\bibinfo{volume}{35}.
  \bibinfo{pages}{32160--32171}.
\newblock


\bibitem[Zhang et~al\mbox{.}(2022)]%
        {ZHANG2022110242}
\bibfield{author}{\bibinfo{person}{Xin Zhang}, \bibinfo{person}{Chao He},
  \bibinfo{person}{Yanping Lu}, \bibinfo{person}{Biao Chen},
  \bibinfo{person}{Le Zhu}, {and} \bibinfo{person}{Li Zhang}.}
  \bibinfo{year}{2022}\natexlab{}.
\newblock \showarticletitle{Fault diagnosis for small samples based on
  attention mechanism}.
\newblock \bibinfo{journal}{\emph{Measurement}}  \bibinfo{volume}{187}
  (\bibinfo{year}{2022}), \bibinfo{pages}{110242}.
\newblock
\showISSN{0263-2241}


\bibitem[Zhao et~al\mbox{.}(2020)]%
        {DRSN}
\bibfield{author}{\bibinfo{person}{Minghang Zhao}, \bibinfo{person}{Shisheng
  Zhong}, \bibinfo{person}{Xuyun Fu}, \bibinfo{person}{Baoping Tang}, {and}
  \bibinfo{person}{Michael Pecht}.} \bibinfo{year}{2020}\natexlab{}.
\newblock \showarticletitle{Deep Residual Shrinkage Networks for Fault
  Diagnosis}.
\newblock \bibinfo{journal}{\emph{{IEEE} Trans. Ind. Informat.}}
  \bibinfo{volume}{16}, \bibinfo{number}{7} (\bibinfo{year}{2020}),
  \bibinfo{pages}{4681--4690}.
\newblock


\bibitem[Zhu et~al\mbox{.}(2023)]%
        {zhu2023spikegpt}
\bibfield{author}{\bibinfo{person}{Rui-Jie Zhu}, \bibinfo{person}{Qihang Zhao},
  \bibinfo{person}{Guoqi Li}, {and} \bibinfo{person}{Jason~K Eshraghian}.}
  \bibinfo{year}{2023}\natexlab{}.
\newblock \showarticletitle{Spikegpt: Generative pre-trained language model
  with spiking neural networks}.
\newblock \bibinfo{journal}{\emph{arXiv preprint arXiv:2302.13939}}
  (\bibinfo{year}{2023}).
\newblock


\bibitem[Zuo et~al\mbox{.}(2022)]%
        {MLSNN}
\bibfield{author}{\bibinfo{person}{Lin Zuo}, \bibinfo{person}{Fengjie Xu},
  \bibinfo{person}{Changhua Zhang}, \bibinfo{person}{Tangfan Xiahou}, {and}
  \bibinfo{person}{Yu Liu}.} \bibinfo{year}{2022}\natexlab{}.
\newblock \showarticletitle{A multi-layer spiking neural network-based approach
  to bearing fault diagnosis}.
\newblock \bibinfo{journal}{\emph{Reliab. Eng. Syst. Saf.}}
  \bibinfo{volume}{225} (\bibinfo{year}{2022}), \bibinfo{pages}{108561}.
\newblock
\showISSN{0951-8320}


\bibitem[Zuo et~al\mbox{.}(2021)]%
        {zuosnn}
\bibfield{author}{\bibinfo{person}{Lin Zuo}, \bibinfo{person}{Lei Zhang},
  \bibinfo{person}{Zhe-Han Zhang}, \bibinfo{person}{Xiao-Ling Luo}, {and}
  \bibinfo{person}{Yu Liu}.} \bibinfo{year}{2021}\natexlab{}.
\newblock \showarticletitle{A spiking neural network-based approach to bearing
  fault diagnosis}.
\newblock \bibinfo{journal}{\emph{J. Manuf. Syst.}}  \bibinfo{volume}{61}
  (\bibinfo{year}{2021}), \bibinfo{pages}{714--724}.
\newblock


\end{thebibliography}

\section*{Appendix}
\appendix

\begin{figure*}[!t]
\centering
\includegraphics[width=6.2in]{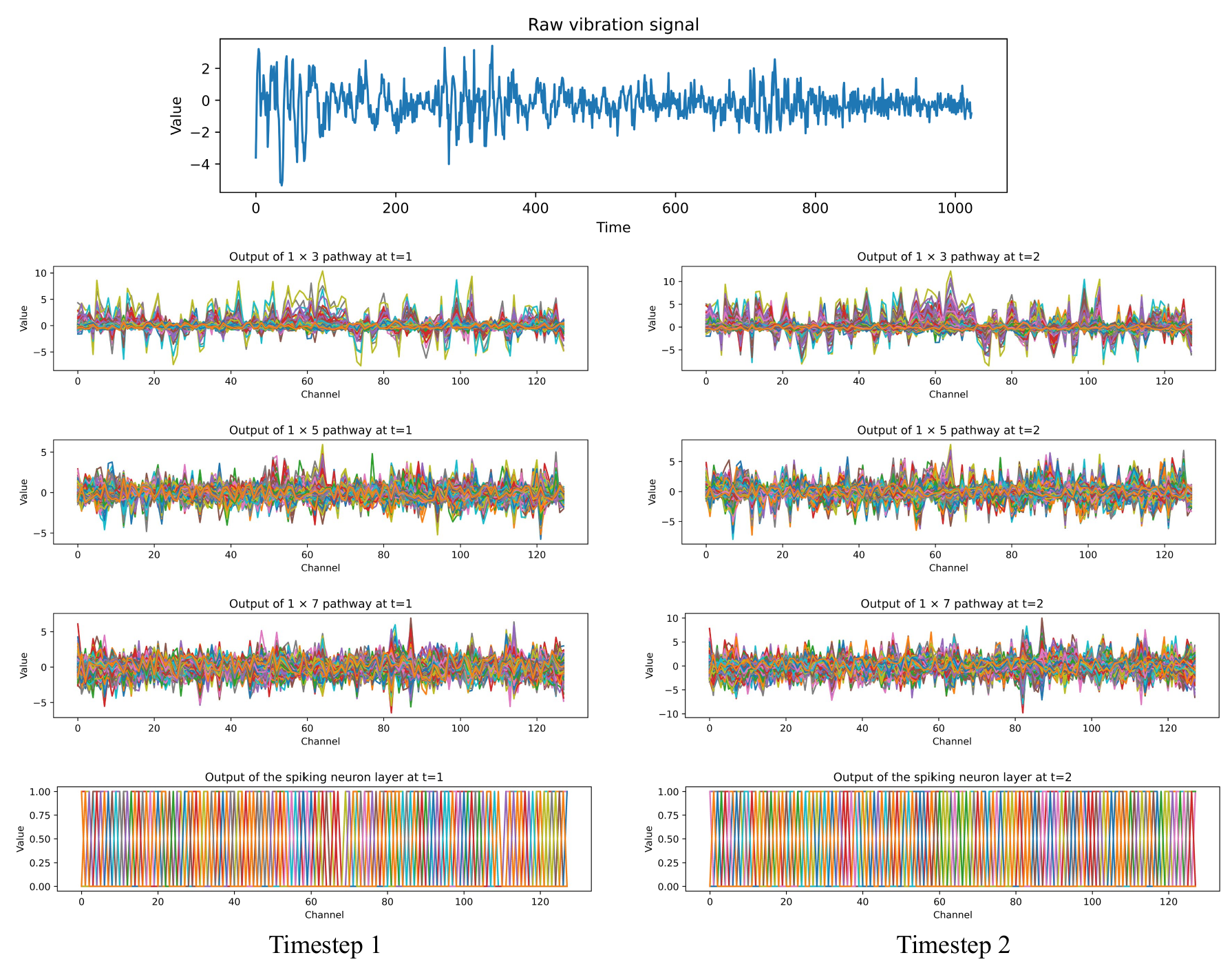}
\vskip -0.12in
\caption{Output visualization of the three convolution pathways and the spiking neuron layer in the multi-scale attention module. From top to bottom are the raw vibration signal, the output of the $1\times3$, $1\times5$, $1\times7$ convolution pathways, and the spiking neuron layer. Each convolution pathway extracts an individual pattern of fault features, and the spiking neurons fuse the input currents to generate discrete 0-1 spikes}
\label{encoding_vis}
\end{figure*}

\section{Multi-Scale Feature Visualization}
\label{msfvis}

To clearly illustrate the effect of the multi-scale attention encoding module, Fig.~\ref{encoding_vis} visualizes the outputs of the three convolution pathways and the spiking neuron layer at the first two timesteps. Each of the three convolution pathways extracted different features related to the fault, and the spiking neurons incorporating the fused input currents generated discrete 0-1 spikes. These 0-1 spikes are passed as input to the follow-on spike residual block for further fault diagnosis.

\section{Experimental Details}
\label{exp_detail}
\subsection{Dataset Description}

\subsubsection{MFPT}
The MFPT~\cite{MFPT} Bearing Fault Dataset is a benchmark dataset for validating bearing fault diagnosis algorithms. The MFPT dataset includes normal, multiple loads outer race, inner race fault bearing data from a bearing test rig, and fault data from three real-world environments. In the experiments, we used data from one baseline condition, seven outer race fault conditions, and seven inner race fault conditions. As a result, a total of 1 normal class and 14 fault classes were generated. Each class contains 140 samples, and each sample vibration signal has a length of 1024, obtained from the raw data using non-overlapping sampling. For evaluating the fault diagnosis model, 70\% of the samples were randomly divided for training the model, and the remaining 30\% of the samples were used for performance evaluation.

\subsubsection{JNU}

The JiangNan University (JNU)~\cite{JNU} bearing fault dataset was collected by Jiangnan University, China. The JNU dataset contains data of four health conditions: (1) normal; (2) outer-race defects; (3) inner-race defects; and (4) roller element defects. Vibration signals with a sampling frequency of 50 kHz were obtained at three rotating speeds, yielding a total of 12 classes. Each class contains 150 samples of length 1024, with 50\% each for training and evaluation.

\subsubsection{Bearing and Gearbox Datasets.} The Bearing and Gearbox datasets~\cite{Gearbox} were provided by Southeastern University and include data from the Driveline Dynamics Simulator. It contains data on normal and four fault types under two operating conditions, and can therefore be divided into a total of 10 categories. We sampled the vibration signals recorded in the data with a length of 1024 in the x, y, and z directions, so the inputs to the model were three channels of signals. Each category contains 200 samples, and 50\% of the samples are randomly divided as the training set for each training.

\begin{table}[!t]
 \centering
 \caption{Structures of ResNet and MRA-SNN. Conv($1 \times 3$) denotes the spiking convolution block, which includes a $1 \times 3$ convolution, a BN layer, and a spiking neuron layer that finally outputs a 0-1 spike sequence. fc denotes the fully connected layer that produces the fault category output.\label{model}}
 \vskip -0.1in
 \begin{tabular}{cc}
  \hline
  ResNet & MRA-SNN \\
  \hline
  Conv($1 \times 3@64$) & -\\
  \hline
  \makecell{
  $\left(
 	    \begin{array}{c}  
 			 \makecell{\text{Conv}(1 \times 3@64) \\\text{Conv}(1 \times 3@64)}
        \end{array}
    \right)\times 2
  $} & 
  \makecell{
  $
   \begin{array}{c}  
 	\text{Conv}(1 \times 3(5,7))@32 \\ \text{Conv}(1 \times 3(5,7))@64
   \end{array}
  $} \\
  \hline
  \makecell{
  $\left(
 	    \begin{array}{c}  
 			 \makecell{\text{Conv}(1 \times 3@128) \\ \text{Conv}(1 \times 3@128)}
        \end{array}
    \right)\times 2
  $} & -\\
  \hline
  \makecell{
  $\left(
 	\begin{array}{c}  
 			 \makecell{\text{Conv}(1 \times 3@256) \\ \text{Conv}(1 \times 3@256)}
 \end{array}
 \right)\times 2
 $} & 
 \makecell{
  $\left(
 	\begin{array}{c}  
 			 \makecell{\text{Conv}(1 \times 3@256) \\ \text{Conv}(1 \times 3@256)}
 \end{array}
 \right)\times 2
 $}\\

  \hline
  \makecell{
  $\left(
 	\begin{array}{c}  
 			 \makecell{\text{Conv}(1 \times 3@512) \\ \text{Conv}(1 \times 3@512)}
 \end{array}
 \right)\times 2
 $} & 
 \makecell{
  $\left(
 	\begin{array}{c}  
 			 \makecell{\text{Conv}(1 \times 3@512) \\ \text{Conv}(1 \times 3@512)}
 \end{array}
 \right)\times 2
 $}\\
  \hline
    \multicolumn{2}{c}{global average pool, fc}\\
  \hline
 \end{tabular}
\end{table}

\subsection{Comparative Models}
For performance comparison, some existing methods are reproduced (For a fair comparison, all methods follow the same training strategy as described above):

$\bullet$ ResNet: The ResNet used for comparison has the same structure as ResNet-18 in the original paper~\cite{ResNet}. To preserve the information in the vibration signal, the first layer uses a $1\times3$ convolution to replace convolution kernel 7 and max pooling, as shown in Table~\ref{model}.

$\bullet$ DRSN: The DRSN~\cite{DRSN} consists of multiple residual shrinkage building units stacked for superior performance and robustness. The DRSN used for comparison has the same structure as the ResNet, but uses channel-wise soft thresholding to eliminate unimportant features.

$\bullet$ Spiking ResNet: Spiking ResNet is the spiking version of ResNet but uses LIF neurons to replace the ReLU activation function in ResNet.

$\bullet$ DSRSN: DSRSN~\cite{DSRSN} is the spiking version of DRSN, with the same structure as DRSN, but using LIF neurons.

$\bullet$ MLR-SNN: MLR-SNN~\cite{MLR-SNN} uses membrane potential learnable LIF neurons. For a fair comparison, MLR-SNN also follows the same structure as ResNet. For the implementation, the extent of leakage is controlled by learnable variables following a sigmoid function.

$\bullet$ MS-ResNet: MS-ResNet~\cite{MSResNet} uses the membrane potential shortcut and is a prototype of an effective SNN architecture. In the implementation, MSResNet has a similar architecture to ResNet to ensure the number of comparable parameters.

$\bullet$ Distillation: We used ResNet-18 as a teacher model for the knowledge distillation of the lightweight ResNet-8, which is based on the~\cite{10496469}.

$\bullet$ DCA-BiGRU: We use the publicly available model architecture and parameters published by~\cite{ZHANG2022110242} for training.

$\bullet$ LiConvFormer: We use the publicly available model architecture and parameters published by~\cite{LiConvFormer} for training.

$\bullet$ Attention SNN: We adopt the same macro-architecture as ResNet, using the attention mechanisms proposed by~\cite{10032591}.

$\bullet$ SE-Net: We adopt the same macro-architecture as ResNet, using the SE attention mechanism proposed by~\cite{SE}.

$\bullet$ CBAM: We adopt the same macro-architecture as ResNet, using the CBAM attention mechanism proposed by~\cite{CBAM}.

\begin{table}[!t]
 \centering
 \caption{Comparison of FLOPs and energy consumption of different models\label{com_energy}}
 \vspace{-0.1in}
 \begin{tabular}{cccc}
  \hline
 Method & MAC & AC & Energy($pJ$)\\
  \hline
  ResNet~\cite{ResNet} & 696,458,752 & 524,288 & 3,204,182,118.4\\
  DRSN~\cite{DRSN} & 698,547,712 & 524,288 & 3,213,791,334.4\\
  LEFE-Net~\cite{LEFE-Net} & 132,467,712 & - & 609,351,475.2\\
  Distillation~\cite{10496469} & 276,827,904 & 262,144 & 1,273,644,288\\
  LiConvFormer~\cite{LiConvFormer} & 51,906,262 & 98,304 & 238,857,278.8\\
  Spiking ResNet & 786,432 & 549,756,242 & 498,398,205.4\\
  DSRSN~\cite{DSRSN} & 9,142,272 & 440,055,495 & 438,104,396.4\\
  MLR-SNN~\cite{MLR-SNN} & 786,432 & 481,511,428 & 436,978,232.1\\
  MS-ResNet~\cite{MSResNet} & 786,432 & 512,848,700 & 516,466,288.8\\
  \hline
  MRA-ANN & 268,023,681 & 131,072 & 1,233,026,897.4\\
  \textbf{MRA-SNN (Ours)} & 5031040 & 25072983 & 45,708,468.7\\
  \hline
 \end{tabular}
\end{table}

\section{Energy Analysis}
\label{EnergyAna}

In neural networks, the number of floating-point operations (FLOPs) is a typical metric used to evaluate the computational burden. For ANNs, their floating point operations are all MAC operations. As for the SNN, all are binary spike features, i.e., AC operations, except for the first layer, where the inputs are floating point values that introduce MAC operations. Similar to~\cite{Kundu_2021_WACV}, we define the layer's spike activity rate (LSAR) as the number of spikes as a proportion of all neurons. Averaging over $T$ timesteps yields the layer's average spike activity rate LASAR. The corresponding LASAR for the Conv and FC layers are $\Phi^{n}_{Conv}$ and $\Phi^{m}_{FC}$, respectively.

\begin{table*}[!t]
 \centering
 \caption{FLOPs analysis of ANN and SNN models with one-dimensional inputs. $i_m$ and $o_m$ are the input and output dimensions of the FC layer, $\Phi^{0}_{Conv}=1$ and $\Phi^{0}_{FC}=\Phi^{N}_{Conv}$, where $N$ is the number of convolutional layers.\label{FLOP}}
 \vspace{-0.1in}
\begin{tabular}{cccc}
\hline \multirow{2}{*}{ Model } & \multicolumn{3}{c}{ FLOPs of a CONV or FC layer } \\
\cline { 2 - 4 } & Variable & Value & FLOP Type \\
\hline \multirow{2}{*}{ANN} &  $F L_{\text {Conv }}^{n}$  &  $k_{n} \cdot h_{n} \cdot c_{n-1} \cdot c_{n}$  & MAC \\
 & $F L_{F C}^{m}$  &  $i_{m} \cdot o_{m}$  & MAC \\
\hline \multirow{2}{*}{SNN} &  $F L_{SNNConv}^{n}$  & $ T \cdot F L_{\text {Conv }}^{n} \cdot \Phi^{n-1}_{Conv}$  & MAC (n=1) or AC (n>1) \\
&  $F L_{S N N F C}^{m}$  & $ T \cdot F L_{FC}^{n} \cdot \Phi_{FC}^{m-1}$  & AC \\
\hline
\end{tabular}
\end{table*}

The number of FLOPs for the ANN and SNN models for one-dimensional inputs is shown in Table~\ref{FLOP}, where the number of FLOPs for the ANN model is calculated based on~\cite{molchanov2017pruning}. When evaluating the energy consumption, we assume that the hardware platform is a 32-bit floating-point implementation in 45nm technology, where the energy consumption of AC and MAC is $0.9pJ$ and $4.6pJ$, respectively~\cite{6757323}

It is worth noting that there are a large number of residual connections in the ResNet architecture that involve AC operations and are not shown in Table~\ref{FLOP}. In addition, there are some attention mechanisms in DSRSN~\cite{DSRSN} and our MRASNN that cause additional MAC operations. In the experimental section of the power comparison, we have accounted for these additional overheads.

\subsection{Energy Comparison}

A comparison of the performance and energy consumption of MRA-SNN and the comparative methods is shown in Table~\ref{com_result}, and the FLOPs and energy consumption of these detailed operations are shown in Table~\ref{com_energy} (The results of the energy consumption analysis were obtained from tests on the MFPT dataset. For the same input length, the power consumption of the ANN remains constant and the power consumption of the SNN varies due to the difference in spike firing rate.). As can be seen from the comparison results, benefiting from the lightweight architecture, our MRA-SNN requires significantly fewer MAC operation compared to its ANN counterparts, and consumes only 1.43\% of the energy of ResNet~\cite{ResNet}. Even compared to lightweight ANN models~\cite{LEFE-Net,10496469,LiConvFormer}, the event-driven MRA-SNN has lower power consumption. Compared to SNN models Spiking ResNet, MLR-SNN~\cite{MLR-SNN}, and MS-ResNet~\cite{MSResNet}, our MRA-SNN has slightly more MAC operations due to the additional attention mechanism, but the AC operations are drastically reduced, requiring only 8.85\% to 10.43\% of the energy consumption of these models. These energy consumption analyses demonstrate the ability of our method to perform fault diagnosis with minimal energy consumption, thus making it more conducive to deployment in real-world industrial scenarios.

In Table~\ref{com_energy}, we also compare MRA-ANN, an ANN with the same architecture as our MRA-SNN. The results show that MRA-ANN consumes 27 times more power than MRA-SNN for the same architecture. This illustrates the inherent low-power advantage of SNNs and further highlights the need to introduce SNNs for deployment in real-world industrial scenarios.

\section{Specific Details About Adding Noise}
\label{addnoise}
\begin{figure}[h]
\centering
\includegraphics[width=3.2in]{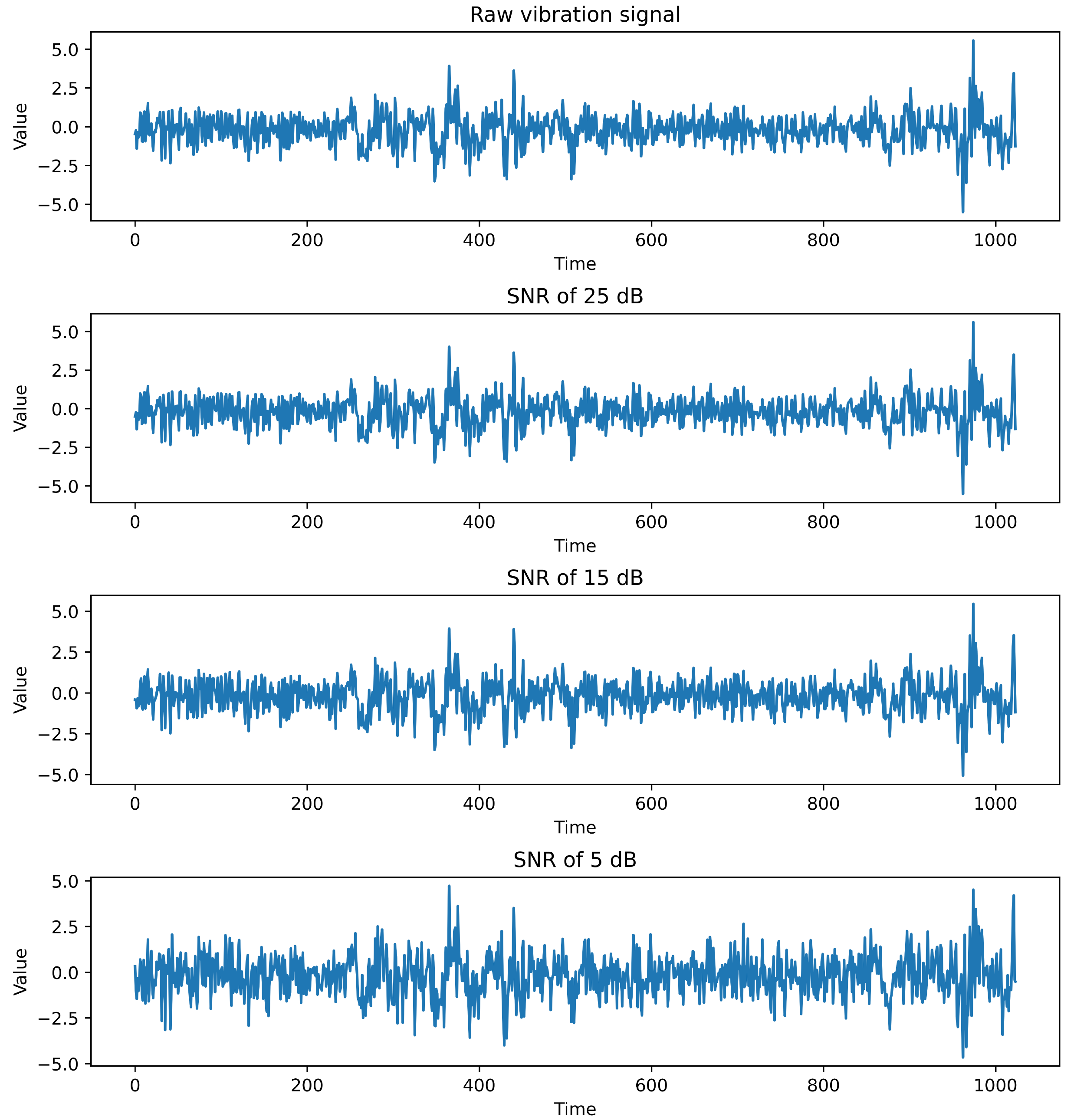}
\vskip -0.1in
\caption{Visualization of the raw vibration signal with noise signals of different SNRs. The lower the SNR, the worse the signal distortion.}
\label{noise_vis}
\end{figure}

Based on the amplitude strength of the original vibration signal, we add different levels of Gaussian noise to it in order to generate noisy interference signals with different SNRs. A lower SNR indicates a greater influence of noise, as shown in Fig.~\ref{noise_vis}. When evaluating the robustness of fault diagnosis methods, the range of SNR is kept from 30 dB to 0 dB.

\begin{figure*}[!t]
\centering
\subfloat[DRSN~\cite{DRSN}]
	{
	\includegraphics[width=2.2in]{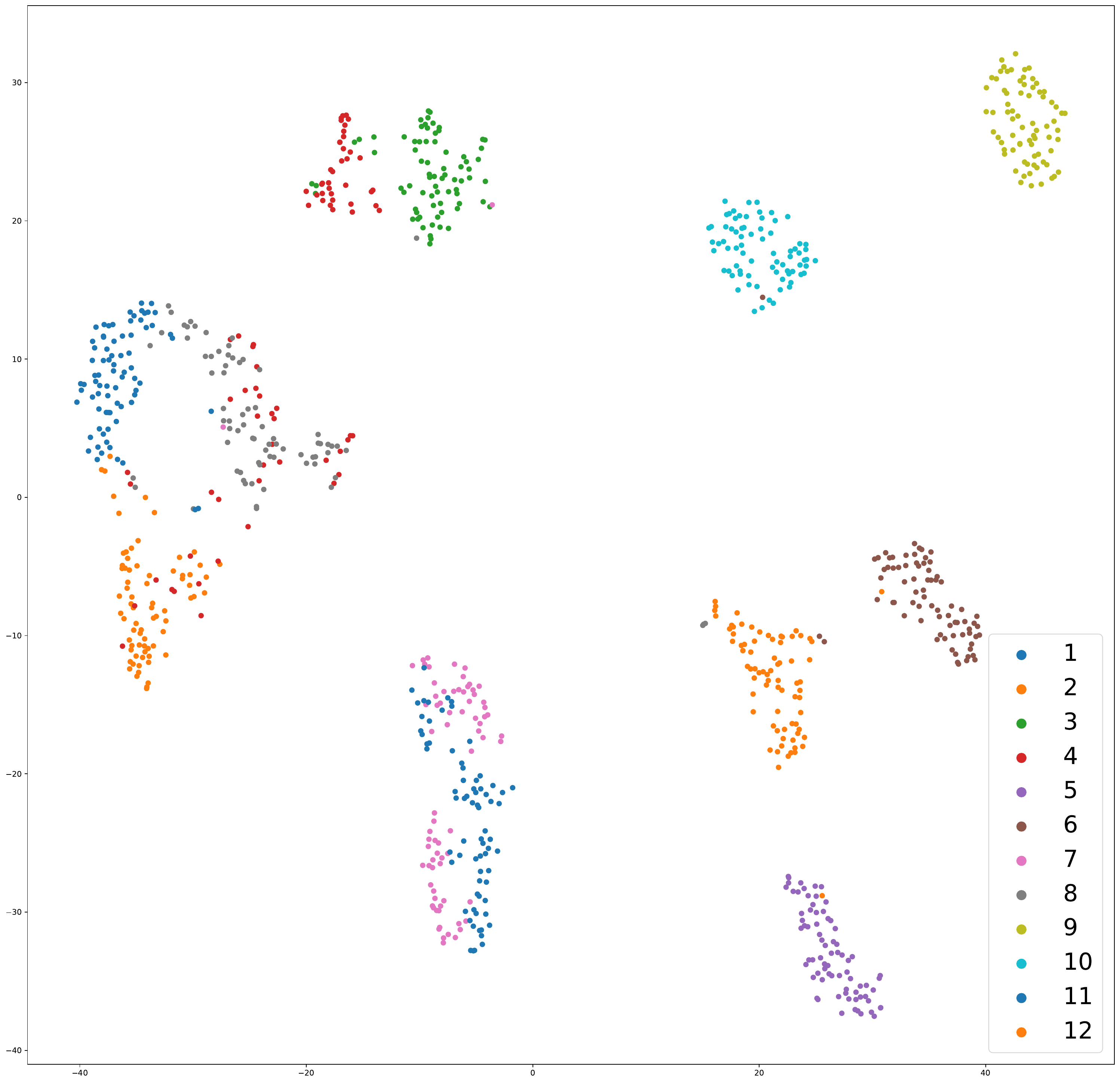}
	}
\subfloat[MLR-SNN~\cite{MLR-SNN}]
	{
	\includegraphics[width=2.2in]{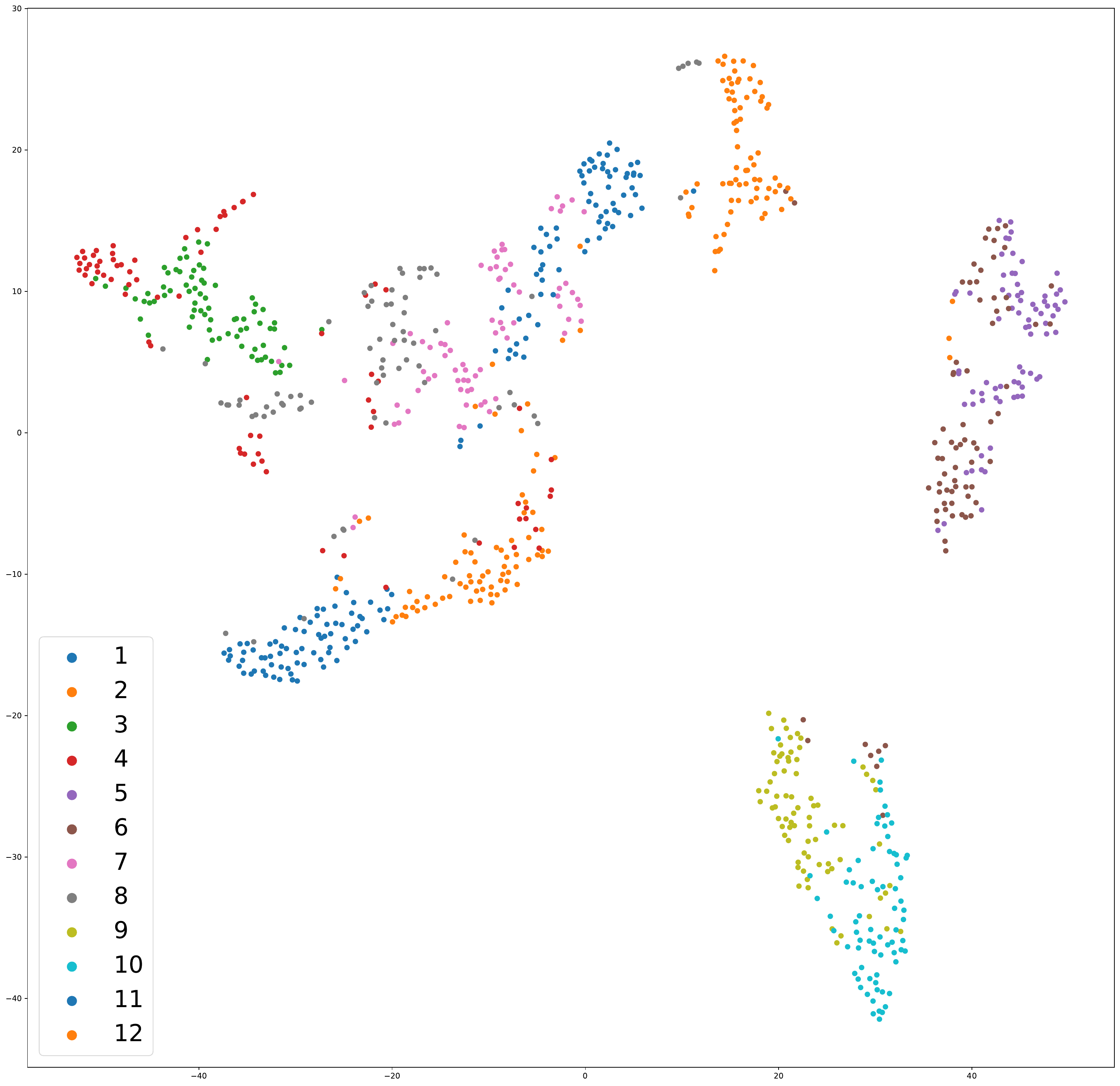}
	}
\subfloat[MRA-SNN (Ours)]
	{
	\includegraphics[width=2.2in]{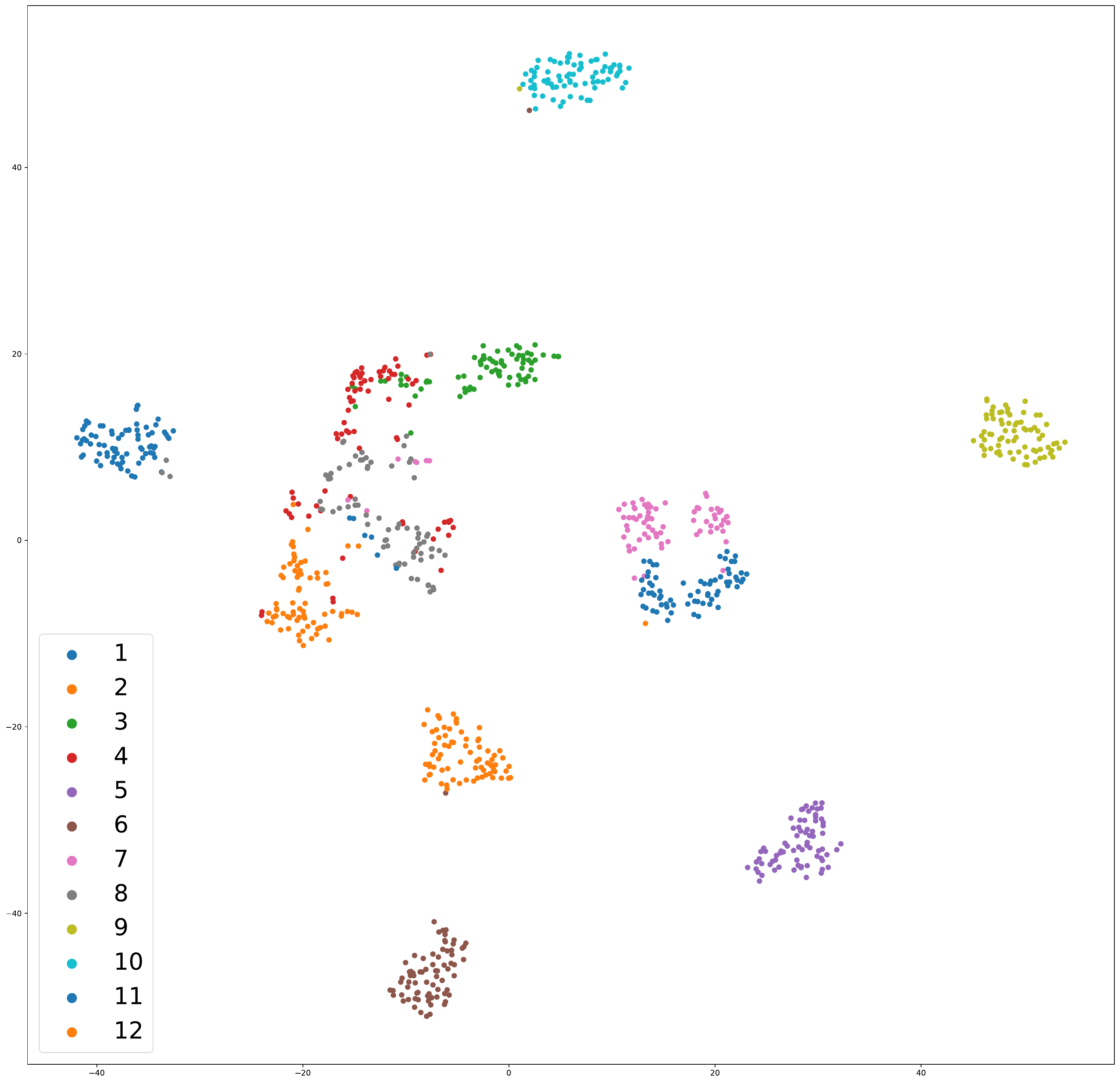}
	}
\caption{Two-dimensional t-SNE visualization on the JNU dataset.}
\label{JNUtsne}
\end{figure*}

\section{Additional visualization Analysis}
\label{AVA}
Fig.~\ref{JNUtsne} shows the t-SNE visualization results on JNU. On the JNU dataset, DRSN~\cite{DRSN} has better classification ability than MLR-SNN~\cite{MLR-SNN}, but there is still severe sticking between multiple faults. The MRA-SNN visualization results show that individual fault clusters are more compact and there is relative separation between different fault clusters. Similar to the MFPT dataset, the MRA-SNN also exhibits superior fault recognition on the JNU dataset. This illustrates the great generalizability of MRA-SNN.

\section{Comparative Results of Confusion Matrices}
\label{com_cm}

\begin{figure*}[!t]
\centering
\subfloat[DRSN~\cite{DRSN}]
	{
	\includegraphics[width=2.2in]{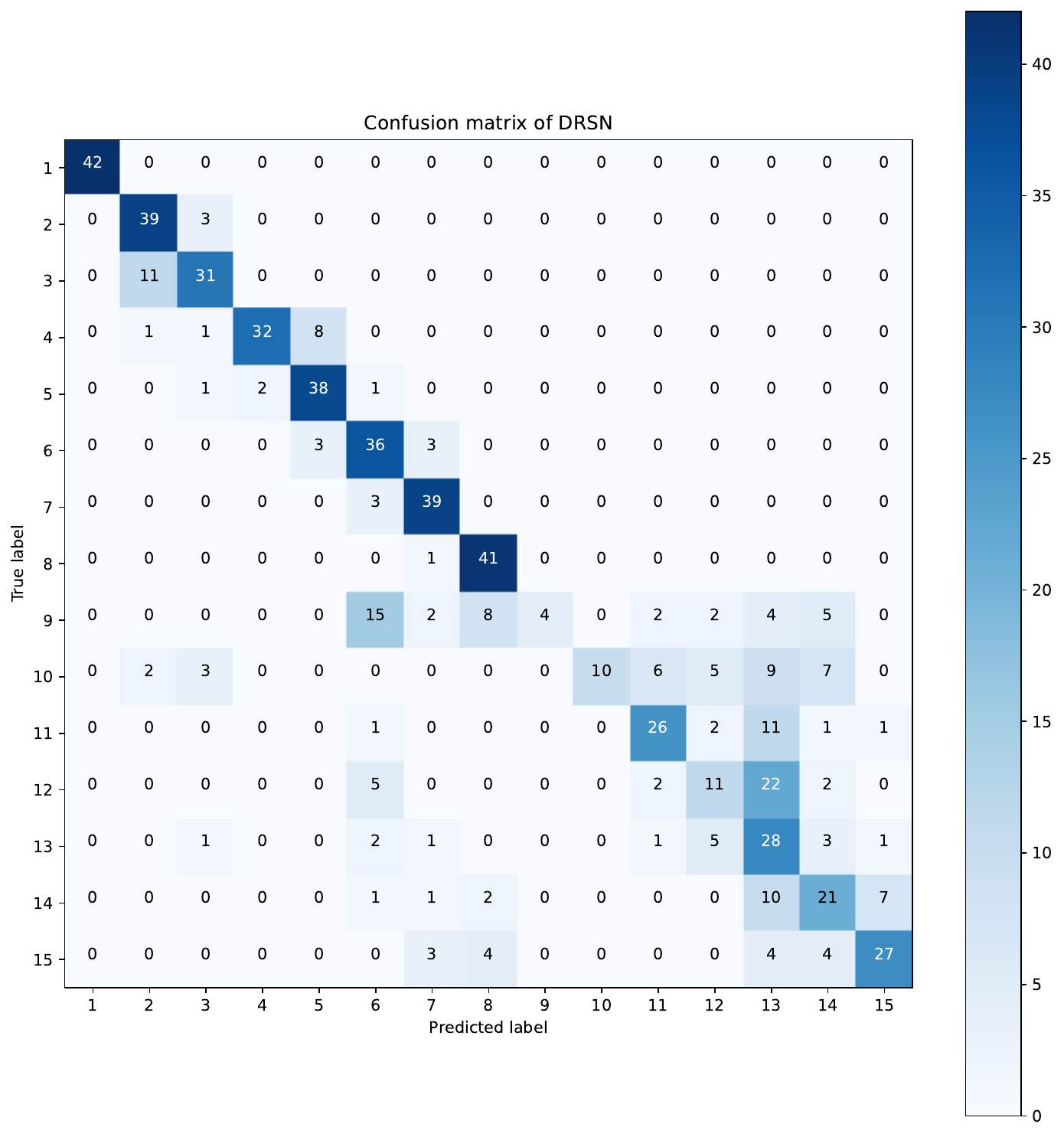}
	}
\subfloat[MLR-SNN~\cite{MLR-SNN}]
	{
	\includegraphics[width=2.2in]{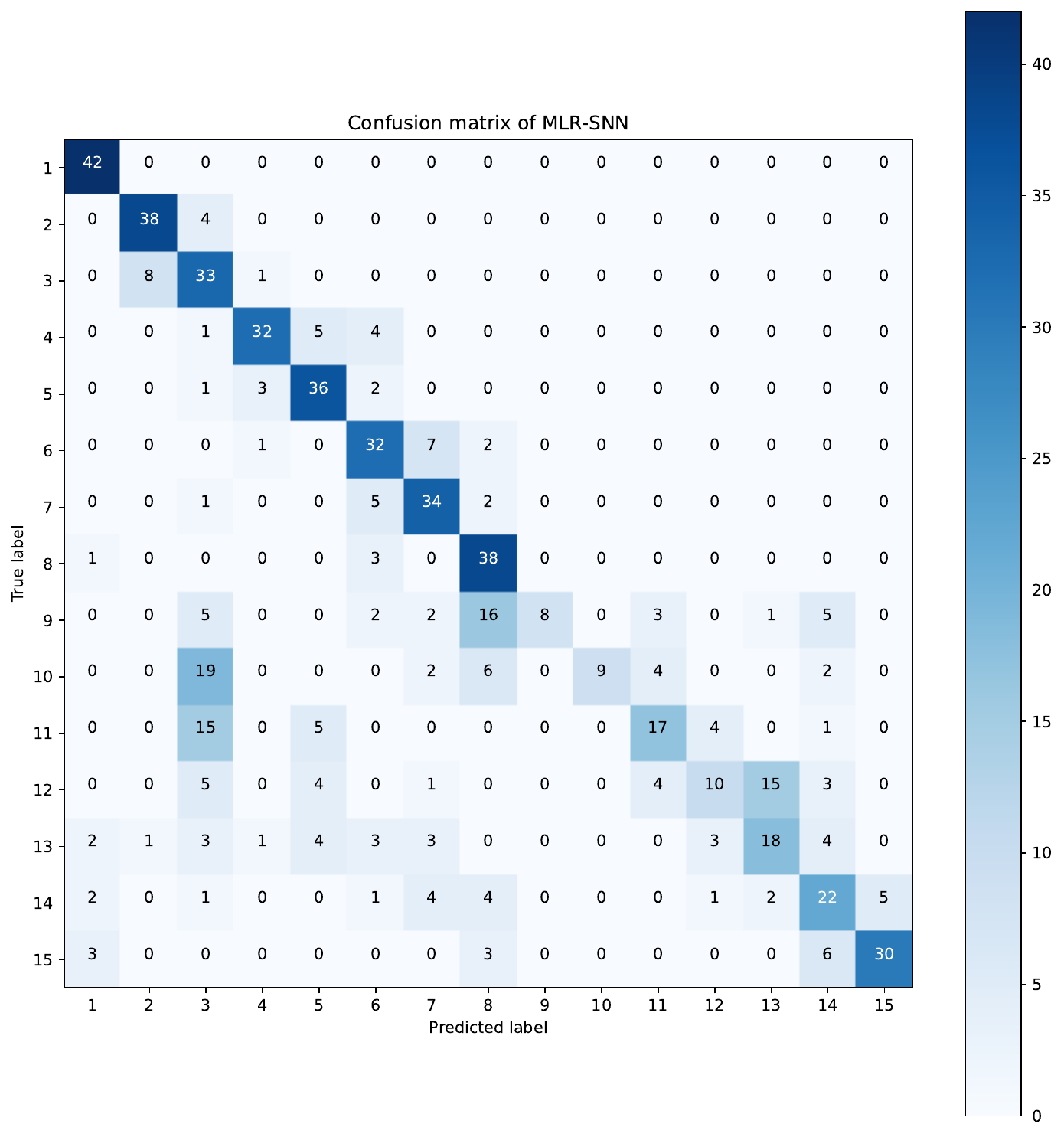}
	}
\subfloat[MRA-SNN (Ours)]
	{
	\includegraphics[width=2.2in]{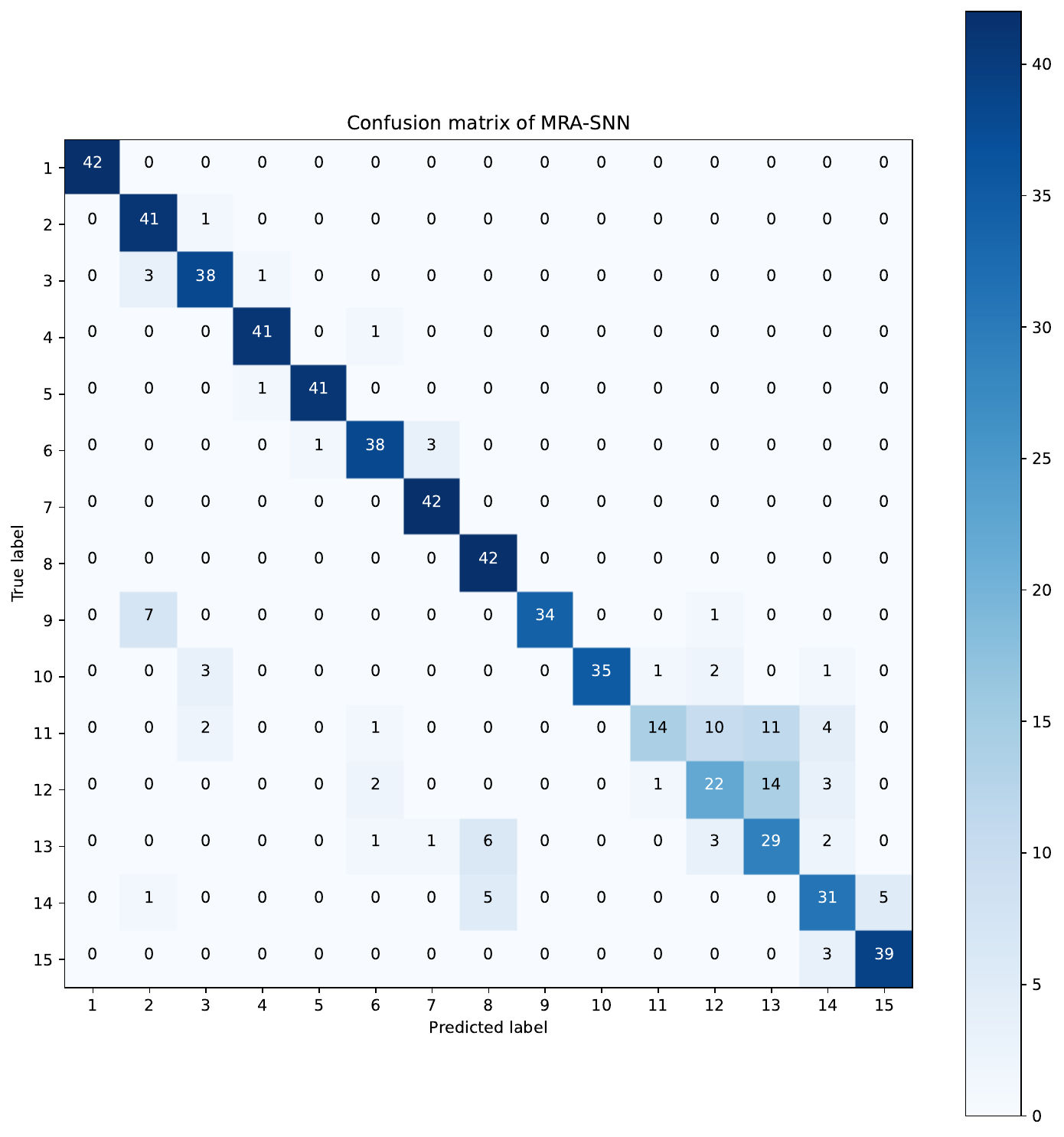}
	}
\caption{Confusion matrix of DRSN, MLR-SNN and proposed MRA-SNN on MFPT for SNR of 10 dB. The overall performance of the MRA-SNN is significantly better than that of the DRSN and the MLR-SNN, especially for faults 9, 10, 14, and 15.}
\label{cm}
\end{figure*}

To further demonstrate the robustness of the MRA-SNN in high-noise environments, Fig.~\ref{cm} visualizes the confusion matrices of the DRSN~\cite{DRSN}, MLR-SNN~\cite{MLR-SNN}, and the proposed MRA-SNN on MFPT at a SNR of 10 dB. As can be seen in Fig.~\ref{cm}, faults 9, 10, and 12 are difficult to recognize for both DRSN and MLR-SNN, and MLR-SNN cannot even recognize faults 11 and 15. In contrast, MRA-SNN is capable of recognizing most of the faults from fault 9 to fault 15. In particular, for fault 15, the accuracy of MRA-SNN exceeds 90\%, while DRSN and MLR-SNN can only reach 64.29\% and 71.43\%, respectively. Although MRA-SNN is slightly less accurate for fault 11, its overall performance substantially exceeds that of the comparative models. These extensive experiments and confusion matrix visualizations confirm the superior noise robustness associated with MRA-SNN.

\begin{figure*}[h]
\centering
\includegraphics[width=6in]{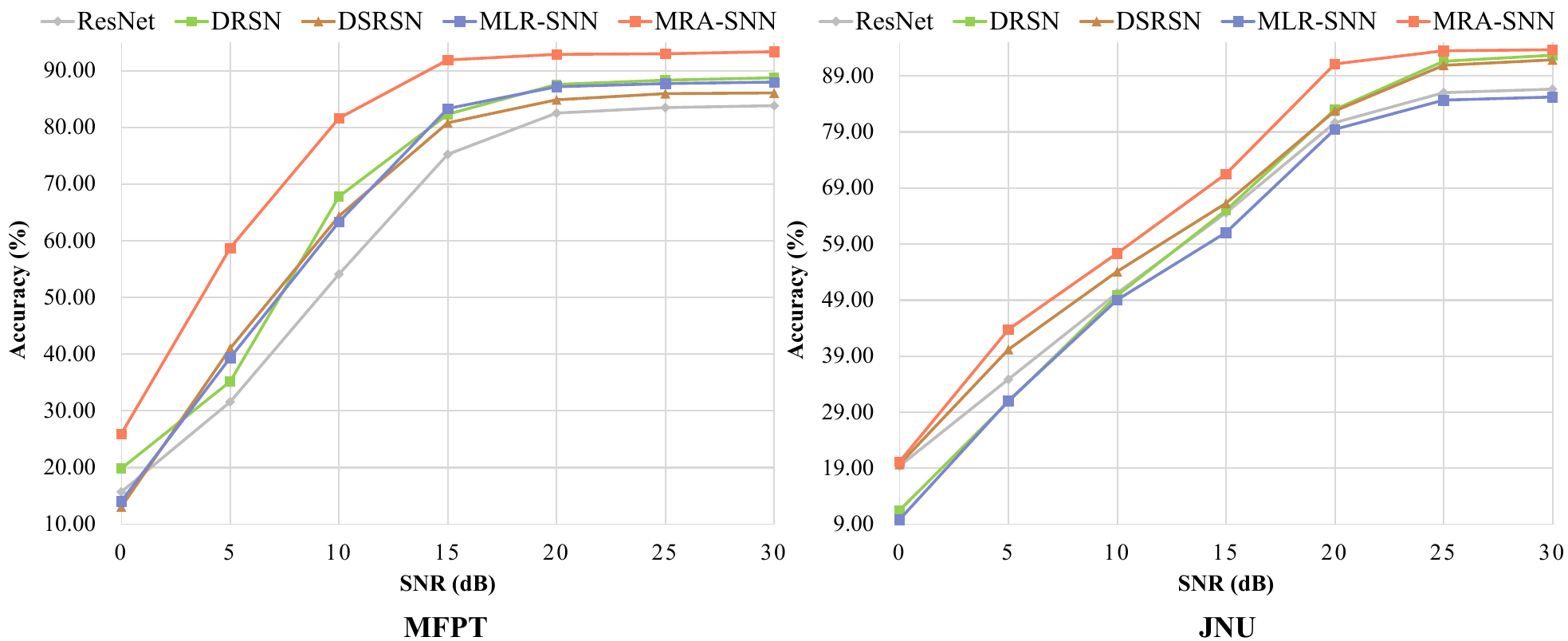}
\caption{Influence of noise on the performance of fault diagnosis models. Compared with other comparative models, our proposed MRA-SNN has consistently higher diagnostic accuracy at any SNR.}
\label{noise_com}
\end{figure*}

The trend of the diagnostic performance of the MRA-SNN and the comparative models with respect to the SNR is shown in Fig.~\ref{noise_com}. Our MRA-SNN consistently shows more accurate diagnostic capability than other models when the SNR ranges from 0 dB to 30 dB. Especially when the noise is relatively mild (15 to 30 dB SNR on the MFPT and 20 to 30 dB on the JNU), there is little degradation in the performance of the MRA-SNN. This demonstrates the superior noise immunity of the MRA-SNN and its practicality in interference-prone mechanical environments.

\end{document}